\documentclass[10pt,twocolumn,letterpaper]{article}

\usepackage[pagenumbers]{cvpr} 

\newcommand{\method}[1]{\textsc{#1}}
\newcommand{\Ours}{\method{Finer-Personalization Rank}\xspace}

\renewcommand{\paragraph}[1]{\vspace{.5em}\noindent\textbf{#1.}}

\usepackage{cuted}
\usepackage{multicol}
\usepackage{multirow}

\definecolor{cvprblue}{rgb}{0.21,0.49,0.74}
\usepackage[pagebackref,breaklinks,colorlinks,allcolors=cvprblue]{hyperref}
\newcommand{\magnify}[0]{\raisebox{-0.005\linewidth}{\includegraphics[width=.035\textwidth]{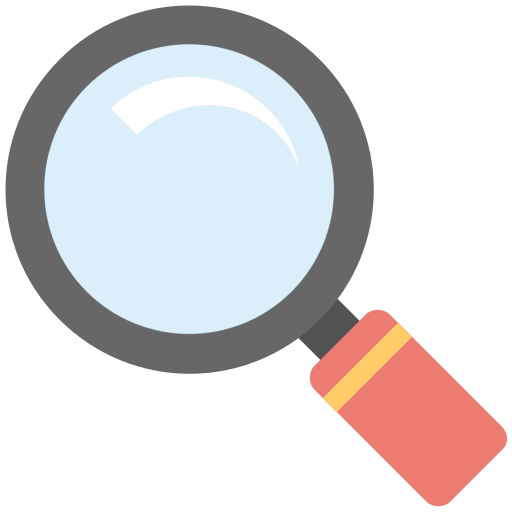}}}

\def\paperID{} 
\def\confName{CVPR}
\def\confYear{2026}

\title{\Ours~\magnify~: Fine-Grained Retrieval\\ Examines Identity Preservation for Personalized Generation}

\author{Connor Kilrain$^{1*}$\quad David Carlyn$^{1*}$\quad Julia Chae$^{2}$\quad Sara Beery$^{2}$\quad Wei-Lun Chao$^{1}$\quad Jianyang Gu$^{1}$\\
{\normalsize $^{1}$ The Ohio State University\quad $^{2}$ MIT}}

\begin{document}
\maketitle

\begin{strip}
\vskip -18pt
\includegraphics[width=\linewidth]{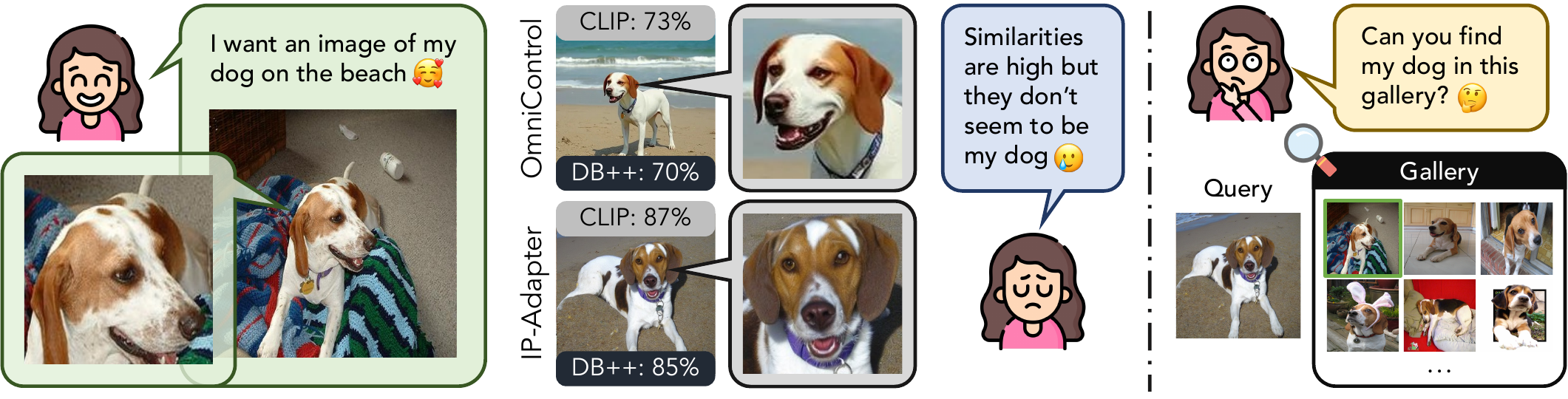}
\vskip -4pt
\captionof{figure}{\textbf{Left}: Given a specific dog as the personalized concept to preserve, two generated images fail to capture the distinctive spot and color patterns on the head. However, popularly adopted evaluation metrics, including CLIP similarity and GPT-based DreamBench++ (DB++)~\cite{peng2025dreambench++}, still give high concept-preserving scores. \textbf{Right}: We propose to evaluate personalization models for their identity preservation capabilities by probing a gallery of visually similar real images with the generated image.}
\label{fig:teaser}
\end{strip}

\begingroup
\renewcommand\thefootnote{}
\footnote{$^{*}$Equal contribution; Source code: \url{https://github.com/OSU-MLB/Finer-Personalization-Rank}}
\addtocounter{footnote}{-1}
\endgroup

\begin{abstract}
The rise of personalized generative models raises a central question: how should we evaluate identity preservation? Given a reference image (\eg, one's pet), we expect the generated image to retain precise details attached to the subject's identity. However, current generative evaluation metrics emphasize the overall semantic similarity between the reference and the output, and overlook these fine-grained discriminative details. We introduce \Ours, an evaluation protocol tailored to identity preservation. Instead of pairwise similarity, \Ours adopts a ranking view: it treats each generated image as a query against an identity-labeled gallery consisting of visually similar real images.
Retrieval metrics (\eg, mean average precision) measure performance, where higher scores indicate that identity-specific details (\eg, a distinctive head spot) are preserved. We assess identity at multiple granularities---from fine-grained categories (\eg, bird species, car models) to individual instances (\eg, re-identification). Across CUB, Stanford Cars, and animal Re-ID benchmarks, \Ours more faithfully reflects identity retention than semantic-only metrics and reveals substantial identity drift in several popular personalization methods. These results position the gallery-based protocol as a principled and practical evaluation for personalized generation.
\end{abstract}    
\section{Introduction}
\label{sec:intro}

Personalized image generation allows users to generate images of \textit{their} dog, \textit{their} favorite plush toy, or \textit{their} personal belongings in different contexts, styles, or scenes~\cite{wei2025personalized}. Recent methods enable this capability by fine-tuning some or all parts of a generative model on a user-provided subject~\cite{ruiz2023dreambooth, kumari2023multi,gal2022image, liu2023cones}. Other works instead inject the user-provided subject into the generation process during inference time~\cite{ye2023ip,li2023blip, tan2025ominicontrol}.
These techniques collectively enable models to capture identity-level visual characteristics and reapply them in novel synthesized contexts. 

As personalized generation becomes widely accessible, it is increasingly important to assess how different models meet user expectations~\cite{peng2025dreambench++,ramesh2022hierarchical}. 
A commonly considered aspect is concept preservation, which measures whether the generated image remains semantically consistent with the reference concept~\cite{ruiz2023dreambooth}.
However, we observe a clear gap between such metrics and practical user demands.
When users use \textit{their} own pets or belongings as the reference concept, they are not merely asking for the correct generic category but for the preservation of instance-specific details that make their subject identifiable. 
We illustrate this discrepancy in~\autoref{fig:teaser}, where the reference beagle has a distinctive spot on the head. The images generated by state-of-the-art personalization models resemble generic beagles but omit this key detail. 
Despite this obvious mismatch to a human observer, existing concept-preservation metrics assign high scores, as they are primarily influenced by overall semantic similarity~\cite{radford2021learning,caron2021emerging}. 
DreamBench++ introduces GPT-based preservation scores that are more aligned with humans~\cite{peng2025dreambench++,hurst2024gpt}, but still struggles to reliably capture subtle discriminative variations. 
This limitation motivates the need for an extended evaluation protocol that more directly tests \textit{identity preservation} rather than coarse semantic concepts to support future progress in personalized AI.  

We argue that in subject-driven personalization, pairwise similarity between a generated image and a single reference is inherently ambiguous. Many different instances can achieve equally high similarity scores, especially when they share a category, pose, or overall background. Such scores provide limited evidence that the model has preserved the specific identity, rather than merely producing another example of the same generic type. To this end, we propose to explicitly introduce visually similar negative examples as a fine-grained gallery. When the generated image is evaluated against a set of near-duplicate candidates, subtle variations in instance-specific details become critical for distinguishing the identity~\cite{schneider2019past,ye2021deep,ye2025transformer}. In this setting, identity preservation is no longer assessed in isolation, but rather through comparative judgments sensitive to fine-grained details. 

We propose \Ours, a gallery-based evaluation protocol built on fine-grained datasets. 
Specifically, given an image as a personalization reference, we use the generated image as a query to probe a large-scale fine-grained gallery consisting of visually similar real images. 
The goal is to retrieve images with the same identity as the reference image. 
The protocol can be performed at multiple granularities---from fine-grained categories (\eg, bird species, car models) to individual instances (\eg, re-identification). 
With finer granularity, the differences between identities are more subtle to locate, demanding stronger identity preservation. By creating contrast with hard negative samples, the protocol can disambiguate identity within either similar or distinct contexts. This setting provides a challenging and realistic testbed for personalized generation, requiring models to preserve fine-grained details beyond generic categorical similarity.

We evaluate existing personalization models on three benchmarks: CUB~\cite{WahCUB_200_2011}, Stanford Cars~\cite{krause2013collecting}, and Animal Re-ID~\cite{adam2025wildlifereid}. 
We find that the introduction of fine-grained galleries exposes more explicit demands of preserving identity-specific details. Additionally, with more focus on such details, specialized models contribute to uncover faithful performance variations across models. Overall, we observe substantial identity drift in many popular personalization models, indicating a gap from practical applications.

\Ours evaluates identity preservation without collecting human preferences or designing prompting strategies. 
The protocol can be feasibly applied to any fine-grained scenarios that meet user requirements. 
It highlights the dimension that current similarity-based metrics often overlook: whether the generated output contains instance-specific details that tell the identity from near-neighbors. This allows us to better surface when personalization methods succeed in the cases users care about.

\section{Related Work}
\label{sec:related}
\subsection{Personalized Generation}
Personalized image generation~\cite{isola2017image, ruiz2023dreambooth, chen2023subject} encapsulates methods that aim to create images jointly conditioned on a user-specified concept while faithfully following a textual prompt. Specifically for our evaluation, we focus on evaluating the subset of methods~\cite{ruiz2023dreambooth,sun2024generative,ye2023ip,labs2025flux,shin2025large,tan2025ominicontrol} capable of preserving the identity of a user-given subject, known as subject-driven generation. Modern efforts are often categorized as fine-tuning or fine-tuning-free approaches. 

Finetuning approaches~\cite{gal2022image,ruiz2023dreambooth,kumari2023multi} learn to generate a given subject by fine-tuning some or all parts of a generation model on a small subset (typically 3-5) of diverse subject-consistent images. Fine-tuning methods often offer stronger identity preservation, yet they require significant time and resources. Moreover, these methods are prone to overfitting and catastrophic forgetting of prior generation capabilities.

Alternatively, non-finetuning approaches~\cite{li2023blip,ye2023ip,tan2025ominicontrol,sun2024generative,shin2025large} instead attempt to address those limitations by providing zero-shot generation via specialized encoders. To accomplish this, these methods extract features from a subject reference image and then inject them into T2I generation. However, these approaches are plagued by their lesser ability to faithfully preserve subject identity in generation.

Within subject-driven image generation, there exist multiple domain-specific approaches that attempt to maximize the preservation of the identity of their specific domain. The most popular specialization is with the preservation of face identity~\cite{xu2024id3,wang2024instantid,xiao2025fastcomposer}
However, data-specific approaches are often limited to their domain and do not translate well to other data domains. As such, we only evaluate subject-driven generation methods that demonstrate general-purpose generation capabilities. 

\begin{figure*}[t]
    \centering
    \includegraphics[width=0.85\textwidth]{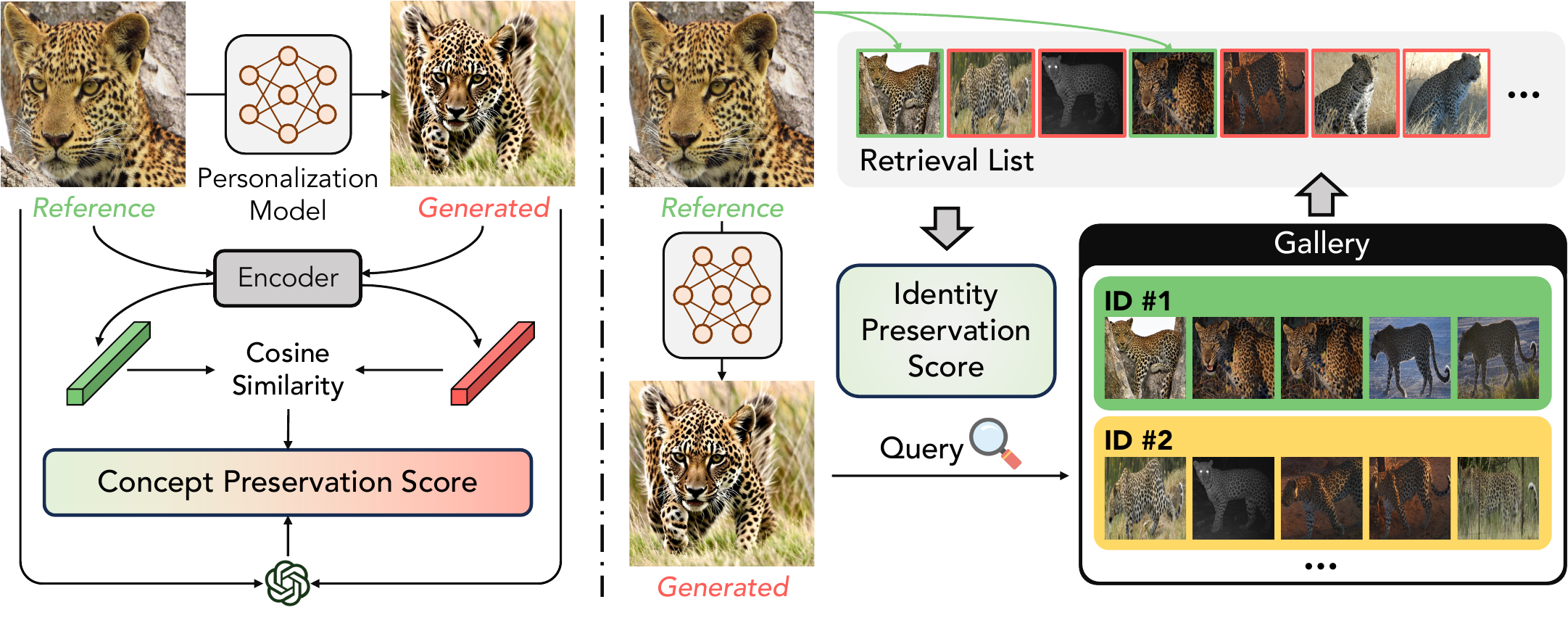}
    \caption{\textbf{Left}: Current evaluation on concept preservation uses general-purpose encoders or GPT to produce a pair-wise similarity score with the reference image. \textbf{Right}: We propose a gallery-based protocol to evaluate identity preservation. The generated image is used to probe a gallery of similar identities. We use mean average precision (mAP) as the measurement. The protocol demands more focus on the retention of fine-grained discriminative details and is more useful in subject-driven personalization scenarios. }
    \label{fig:framework}
\end{figure*}

\subsection{Evaluation Metrics for Generative Models}
There exists a plethora of metrics used to evaluate generative image models. FID~\cite{heusel2017gans} is commonly used to measure the general quality of generated images. CLIP-T~\cite{radford2021learning} is often used to measure the prompt adherence capabilities of generative image models, describing the cosine similarity between the prompt and the generated image CLIP embeddings. LPIPS~\cite{zhang2018unreasonable}, DreamSim~\cite{fu2023dreamsim}, CLIP-I~\cite{radford2021learning}, and DINO~\cite{caron2021emerging} score are used to measure some aspect of perceptual similarity. For measuring subject preservation specifically, CLIP-I and DINO are the two most commonly used metrics. Specifically, both measure the cosine similarity between a subject's reference and generated image embedding vectors using CLIP and DINO to extract image embeddings, respectively. An obvious limitation of doing this is that neither CLIP nor DINO is aligned with fine-grained subject detail, often capturing the more general semantics, shape, and color of an image rather than the necessary identity-defining features. Dreambench++~\cite{peng2025dreambench++} attempts to use GPT-4o~\cite{hurst2024gpt} to provide a more human-aligned measure of subject identity preservation. However, the use of GPT or other SOTA LLMs can cause cost to be a limitation, preventing frequent or intermediate evaluations. A limitation of all prior metrics is their inability to discern the truly important fine-grained details that differentiate identity.

\section{\Ours}
Concept preservation is an important property of personalized generation models. People use \textit{their} pets and belongings as references and prompt personalization models to generate images with customized backgrounds or styles. The expectation is that the generated image accurately retains the distinguishable characteristics of the reference. 

Existing evaluations of concept preservation typically rely on pair-wise similarity scores computed in a general-purpose embedding space (\eg, CLIP~\cite{radford2021learning}) or by LLM-based judgments~\cite{peng2025dreambench++}. These pair-wise similarities usually require only one reference image and are convenient to calculate. However, they primarily capture coarse semantic alignment and are not explicitly designed to resolve instance-level differences between highly similar individuals. 
As a result, they can assign high scores to generations that match the reference in overall layout or appearance, but fail to judge based on the defining details of the reference identity. 
This limits a fair comparison of different personalization models, especially in subject-driven scenarios. 

To address the limitations of similarity-based methods, we extend the evaluation of concept preservation to \textit{identity preservation} and introduce a gallery-based evaluation protocol. Rather than asking whether a generated image is similar to the reference, we ask whether it can be used to correctly retrieve images that match its identity from a gallery of highly similar images. 

\subsection{Retrieval Instead of Similarity}\label{sec:retrieval}

Our framework, shown in~\autoref{fig:framework}, mainly consists of a reference set and a fine-grained gallery, both with real images. 
The reference set is used to prompt personalization. The fine-grained gallery contains images of the same identities that are non-overlapping with the reference set. We start from a randomly allocated gallery from the original dataset, and manually filter out duplicate photos or those with bad angles. We follow the principle of maintaining diversity to reduce the influence of style bias (\eg, similar poses) when manually examining the gallery sets.

For each identity in the reference set, we prompt a given personalization model to generate customized images. 
Image generation is conditioned on diverse prompts to simulate realistic usage, including varying backgrounds (beach, mountain, \etc), poses (lying down, perched, \etc), and actions (running, walking, \etc).
After generating the images, we extract embeddings using a fixed evaluation encoder and use them as queries to probe the gallery. The performance is summarized with mean average precision (mAP). 

Compared with pair-wise similarity, a retrieval-based protocol introduces visually similar objects as negative samples. 
In this case, many different identities may appear similar in the embedding space. 
The similarity score between any two images might be equally high. Instead, we use ranking as a metric to check whether the correct identity outranks other near-neighbors and provide a grounded answer to the question of ``how similar.'' This evaluation well aligns with user intents: to generate images of \textit{their} pets and belongings rather than generically similar ones. 

In addition to direct measurement of identity preservation, adopting fine-grained retrieval as an assessment also eliminates reliance on LLMs or prompt engineering. Once the evaluation encoder and galleries are fixed, the protocol is deterministic, reproducible, and cost-efficient.

\paragraph{Note} This protocol is intended as a complementary measurement of identity preservation for subject-driven generation tasks. It does not replace metrics for instruction following or general image quality, but complements them with information that similarity-based metrics often miss. 

\subsection{Dataset Granularity}
As the protocol requires nearly no extra annotations beyond identity labels, it can be feasibly applied across datasets with different granularities. It allows for flexible evaluations on diverse scenarios that users care about. 
Specifically, in this work, we adopt two complementary granularity levels, including fine-grained categories and individual instances. 
At the fine-grained category level (\eg, species, model, product type~\cite{WahCUB_200_2011,krause2013collecting}), the gallery admits broader intra-category variability. The evaluation requires the generated images to capture distinctive features shared by the category while allowing for other appearance variations. 
By contrast, retrieving the same individual instance (re-identification, Re-ID~\cite{ye2025transformer,ye2021deep,oh2016deep}) more rigorously challenges the identity preservation capabilities of a generative model. The gallery for Re-ID datasets is populated by near-duplicates that share most global attributes. Therefore, successful preservation of the identity requires capturing subtle, localized details (\eg, markings, scars, part shapes) rather than merely category-level features. 

These differences also affect how results should be interpreted. Strong performance at the category level indicates that a method preserves category-related structure across appearance variation, but it cannot tell if the correct individual is retained. Strong Re-ID performance indicates sensitivity to instance-specific details, but cannot reflect if the generated images maintain category-level variations. Using both levels provides complementary views for evaluating personalization models. 

\subsection{Specialized Models for Evaluation}\label{sec:specialized-model}
The gallery-based protocol compares the generated images with many visually similar samples to better reveal identity preservation. But the choice of which encoder is used when ranking similarity has a significant impact on the quality of the evaluation protocol.
General-purpose vision encoders provide robust semantic understanding but are not optimized to resolve instance-level distinctions. 
As identity-specific details are irrelevant to category differentiation, general-purpose models often overlook these details. 
To better measure identity preservation in the retrieval-based protocol, we further consider encoders specialized for fine-grained retrieval and Re-ID. 
The retrieval objectives guide specialized models toward discriminative local features with hard negatives. 
For each fine-grained dataset, we use a specialized model, which is used purely as a frozen encoder. We extract embeddings with the encoder and compute similarity in the corresponding space for ranking.

\section{Experiments}\label{sec:experiments}

\begin{table*}[t]
\centering
\small
\caption{\textbf{Oracle model performances}. The similarity scores are calculated between real images, and mAP scores are calculated with real images as queries to probe the gallery. DB++ indicates the GPT-based concept preservation score in DreamBench++~\cite{peng2025dreambench++}. We only use each specialized model for the corresponding dataset. Therefore, the cross-evaluation on other datasets is skipped (marked as ``-''). ``SP-Cars'' refers to the specialized model trained on Stanford Cars, and ``Sim'' stands for Similarity.}
\label{tab:oracle}
\begin{tabular}{l|ccccc|ccccc|c}
\toprule
\multirow{2}{*}{\textbf{Dataset}} & \textbf{CLIP} & \textbf{DINO} & \textbf{MiewID} & \textbf{BioCLIP} & \textbf{SP-Cars} & \textbf{CLIP} & \textbf{DINO} & \textbf{MiewID} & \textbf{BioCLIP} & \textbf{SP-Cars} & \multirow{2}{*}{\textbf{DB++}} \\
& \textbf{Sim}   & \textbf{Sim}  & \textbf{Sim}    & \textbf{Sim}     & \textbf{Sim} & \textbf{mAP}   & \textbf{mAP}  & \textbf{mAP}    & \textbf{mAP}     & \textbf{mAP} & \\
\midrule
Re-ID & $0.925$ & $0.843$ & $0.647$ & -     & -     & $0.485$ & $0.516$ & $0.803$ & -     & -     & $0.889$ \\
CUB  & $0.837$ & $0.726$ & -     & $0.930$ & -     & $0.520$ & $0.674$ & -     & $0.842$ & -     & $0.748$ \\
Cars  & $0.782$ & $0.434$ & -     & -     & $0.792$ & $0.396$ & $0.228$ & -     & -     & $0.839$ & $0.717$ \\
\bottomrule
\end{tabular}
\end{table*}


\subsection{Experiment Setup}

\begin{figure*}[t]
    \centering
    \includegraphics[width=0.32\textwidth]{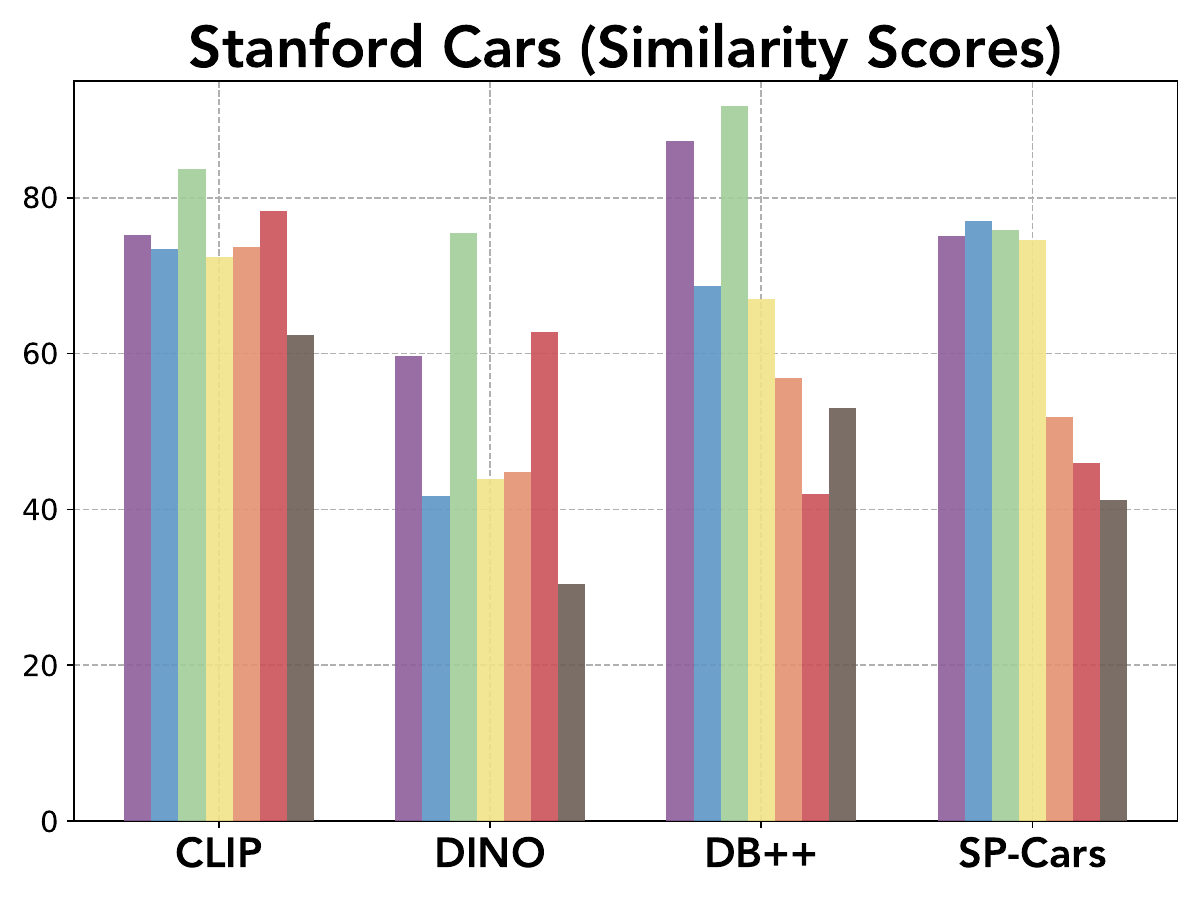}
    \includegraphics[width=0.32\textwidth]{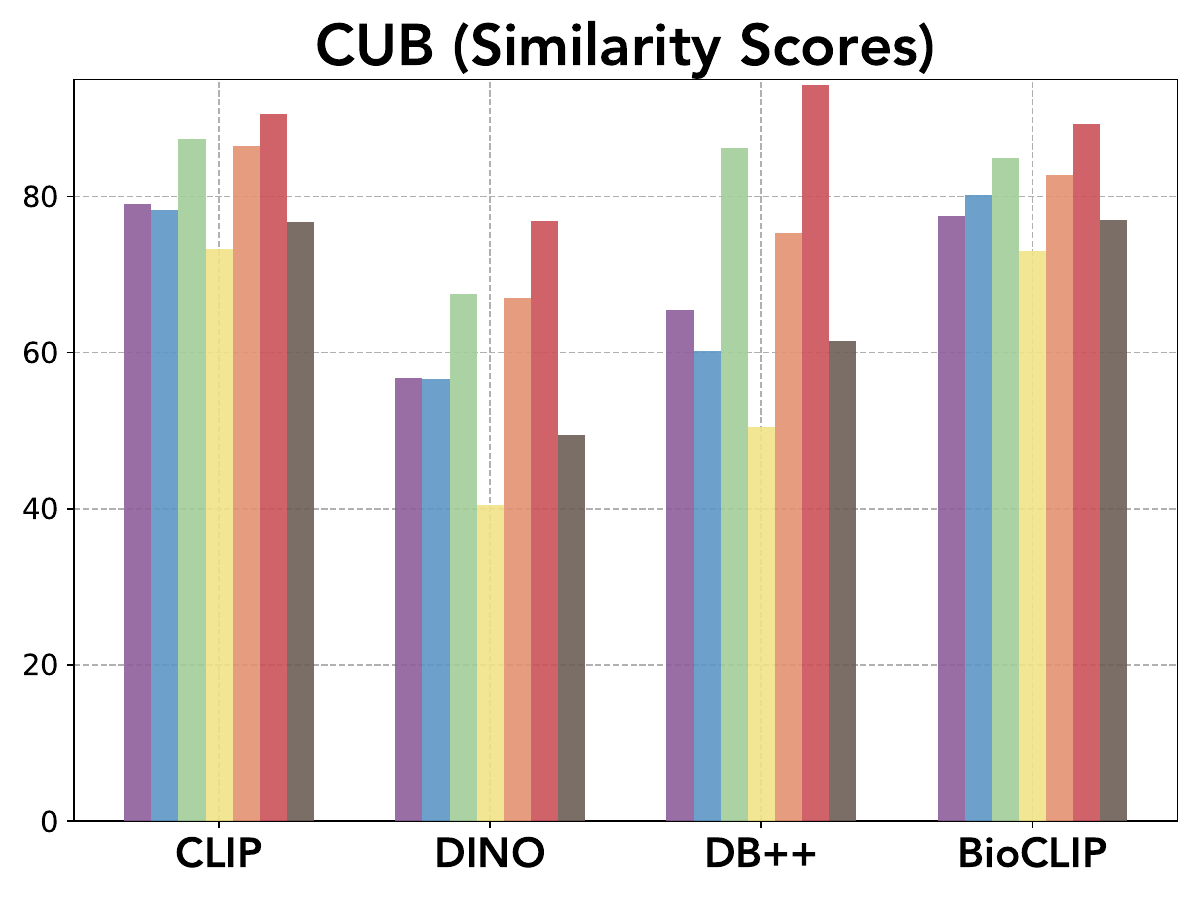}
    \includegraphics[width=0.32\textwidth]{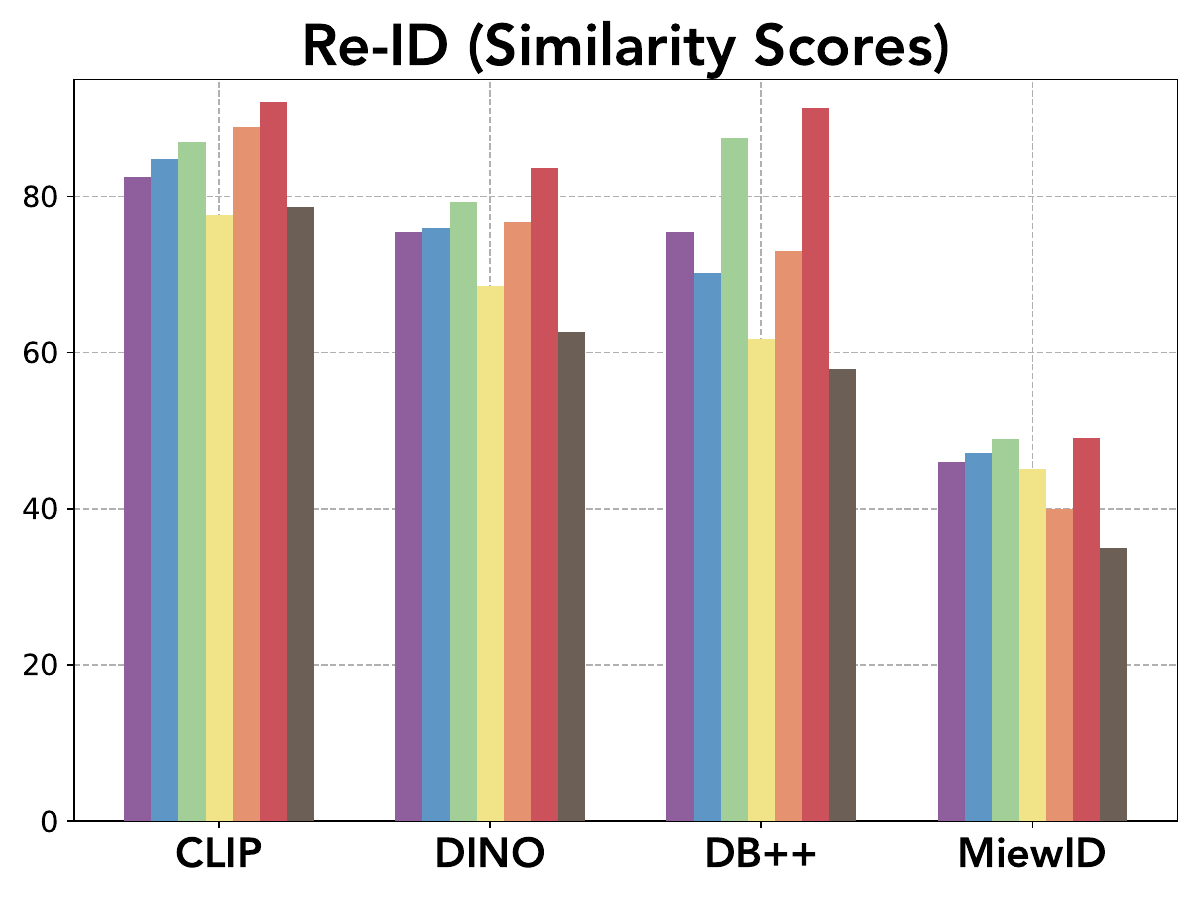}
    \includegraphics[width=0.32\textwidth]{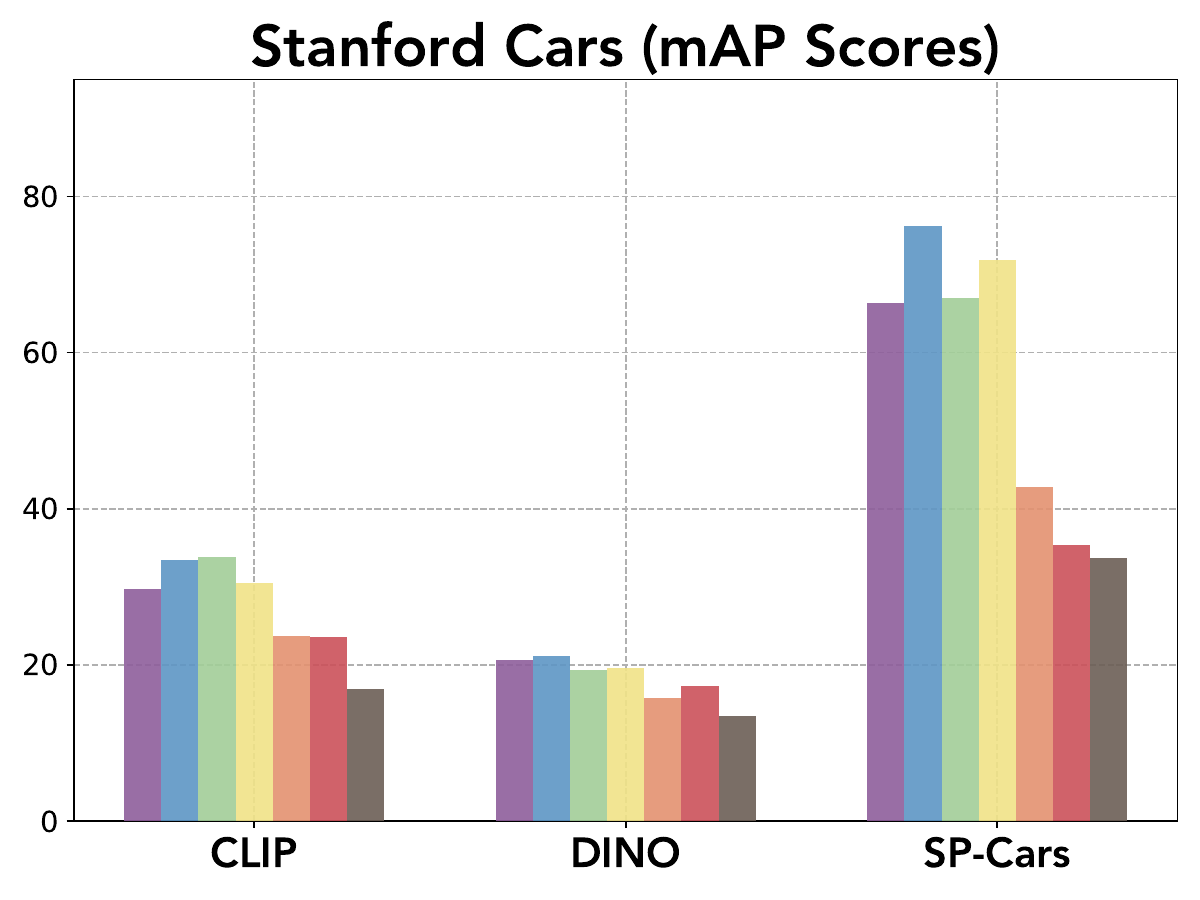}
    \includegraphics[width=0.32\textwidth]{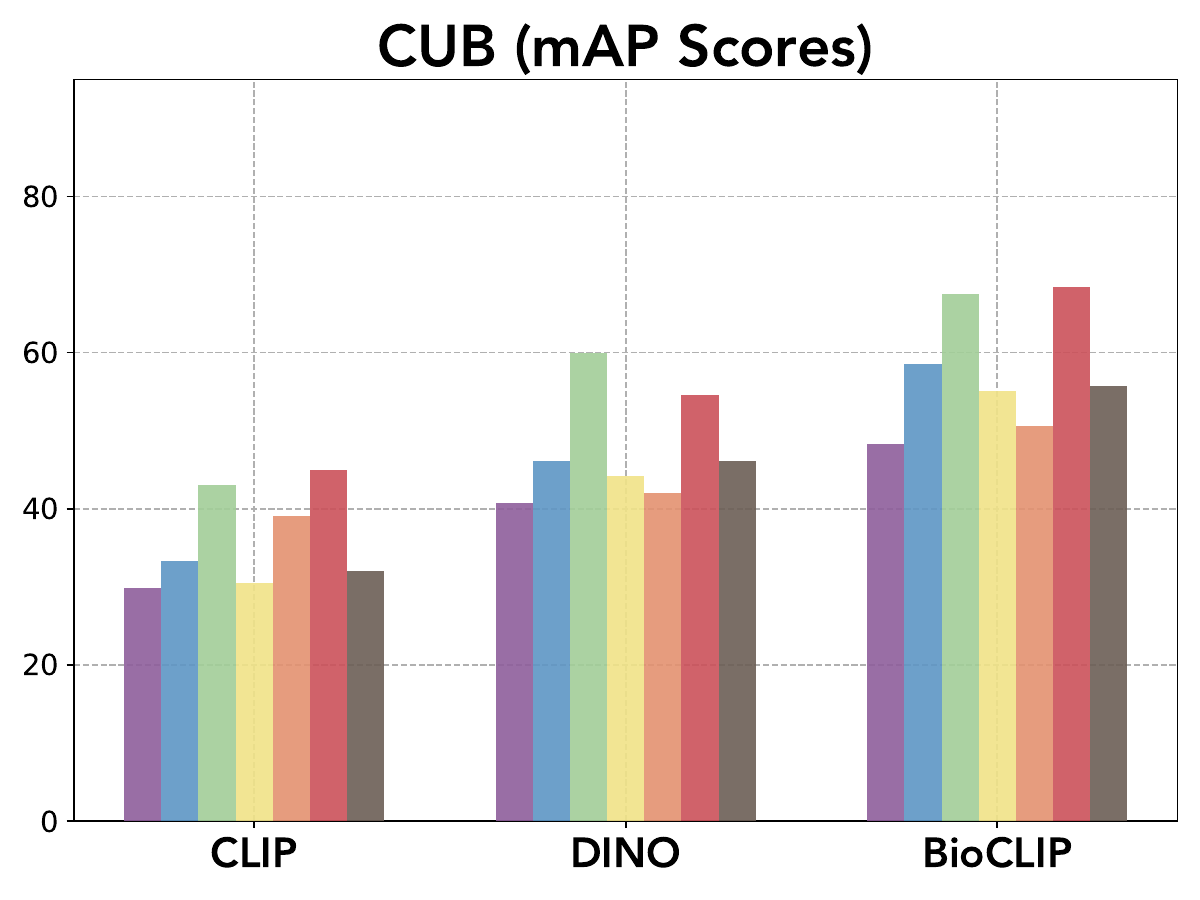}
    \includegraphics[width=0.32\textwidth]{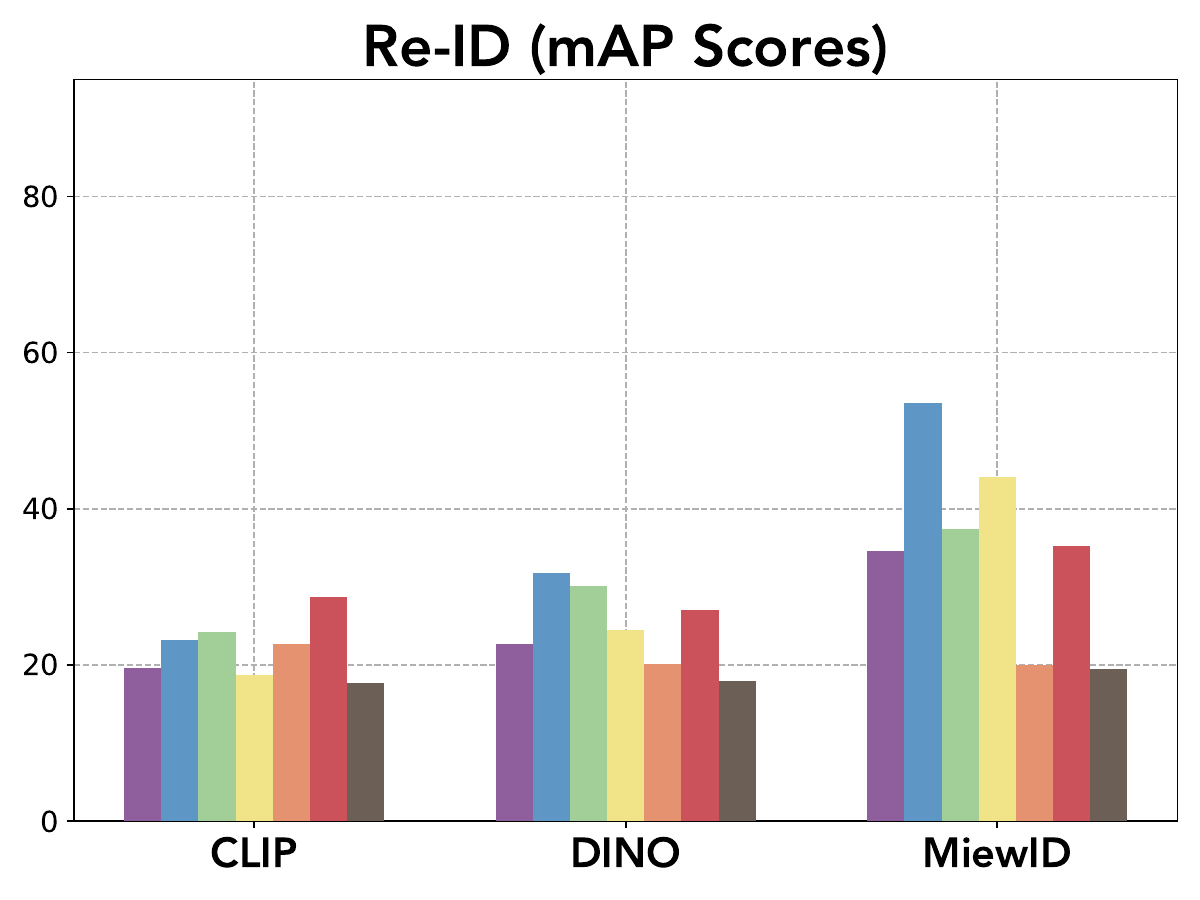}
    \includegraphics[width=0.9\textwidth]{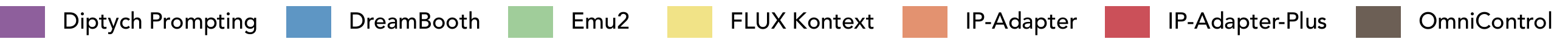}
    \caption{\textbf{Pair-wise similarity scores} (the first row) \textbf{and gallery-based mAP scores} (the second row) of seven personalization models across three benchmarks. The proposed gallery-based protocol more clearly reveals identity drift of these models compared with the overall high similarity scores. Meanwhile, specialized models provide more faithful views to examine identity-related discriminative details.}
    \label{fig:main-result}
    \vspace{-4pt}
\end{figure*}

\paragraph{Datasets}
We employ 3 datasets for our evaluations, including two fine-grained classification datasets and a benchmark suite of animal re-identification datasets across five species. Specifically:
\begin{itemize}
    \item \textbf{CUB}~\cite{WahCUB_200_2011} is a fine-grained dataset for birds. We allocate three groups of species: Wern, Sparrow, and Tern. Each group consists of five visually similar species (\eg, Baird Sparrow, Black Throated Sparrow, Brewer Sparrow, \etc). The group name (\eg, Sparrow) is used in the prompt for generation, instead of the exact species name. 
    \item \textbf{Stanford Cars}~\cite{krause2013collecting} is composed of fine-grained car models. We select 15 visually similar luxury sedan car models from the dataset. When generating images, we use ``car'' in the textual prompt. 
    \item \textbf{WildlifeReID-10k}~\cite{adam2025wildlifereid} is a collection of animal Re-ID datasets. We select a subset of data from this benchmark covering five animal species with distinct biometric appearance markers like spots and stripes (cow, zebra, tiger, giraffe, leopard), sampling ten unique identities for each species. We use the common name in the image generation prompts.
\end{itemize}
For all the evaluations, we separate the dataset into a reference set and a gallery set. Each reference set contains 5-10 images per subject, and each gallery has 10 subjects, each with approximately 10 images.
We use the principles listed in \autoref{sec:retrieval} to construct gallery sets and ablate the building choices in \autoref{sec:abl-gallery}. 
The reference set is exclusively used as a subject reference to prompt the generation. The generated images are then used to probe the gallery set, which is unseen during generation. 
This approach diverges from prior works where a single set of images was used interchangeably for reference and evaluation. The unseen gallery data contributes to a more rigorous evaluation for personalization models. 

\noindent \textbf{Models}
We conducted our evaluations on 7 relevant personalization methods, including \textbf{DreamBooth}~\cite{ruiz2023dreambooth}, \textbf{OminiControl}~\cite{tan2025ominicontrol}, \textbf{Emu2}~\cite{sun2024generative}, \textbf{Diptych Prompting}~\cite{shin2025large},  \textbf{FLUX Kontext}~\cite{labs2025flux}, \textbf{IP-Adapter ViT-G}~\cite{ye2023ip}, and \textbf{IP-Adapter-Plus ViT-H}~\cite{ye2023ip}. Relevant base T2I models were used with each method: DreamBooth with SD 3.0 Medium~\cite{esser2024scaling}, and IP-Adapter methods with SDXL 1.0~\cite{podell2023sdxl}. 
We use OpenAI ViT-B/32 CLIP~\cite{radford2021learning} and DINOv2~\cite{oquab2024dinov} as general-purpose visual encoders.

\noindent \textbf{Specialized models}
As illustrated in~\autoref{sec:specialized-model}, specialized models often have better capabilities in capturing distinctive details of identities. 
We use MiewID~\cite{otarashvili2024multispecies} for Re-ID datasets, BioCLIP 2~\cite{stevens2024bioclip,gu2025bioclip} for CUB, and an EfficientNetv2 model~\cite{tan2021efficientnetv2} trained on the whole Cars training set.
For simplicity, we use SP-Cars, BioCLIP, CLIP, and DINO to represent the specialized model for Cars, BioCLIP 2, OpenAI CLIP, and DINOv2, respectively.

\begin{figure*}[t]
    \centering
    \includegraphics[width=\textwidth]{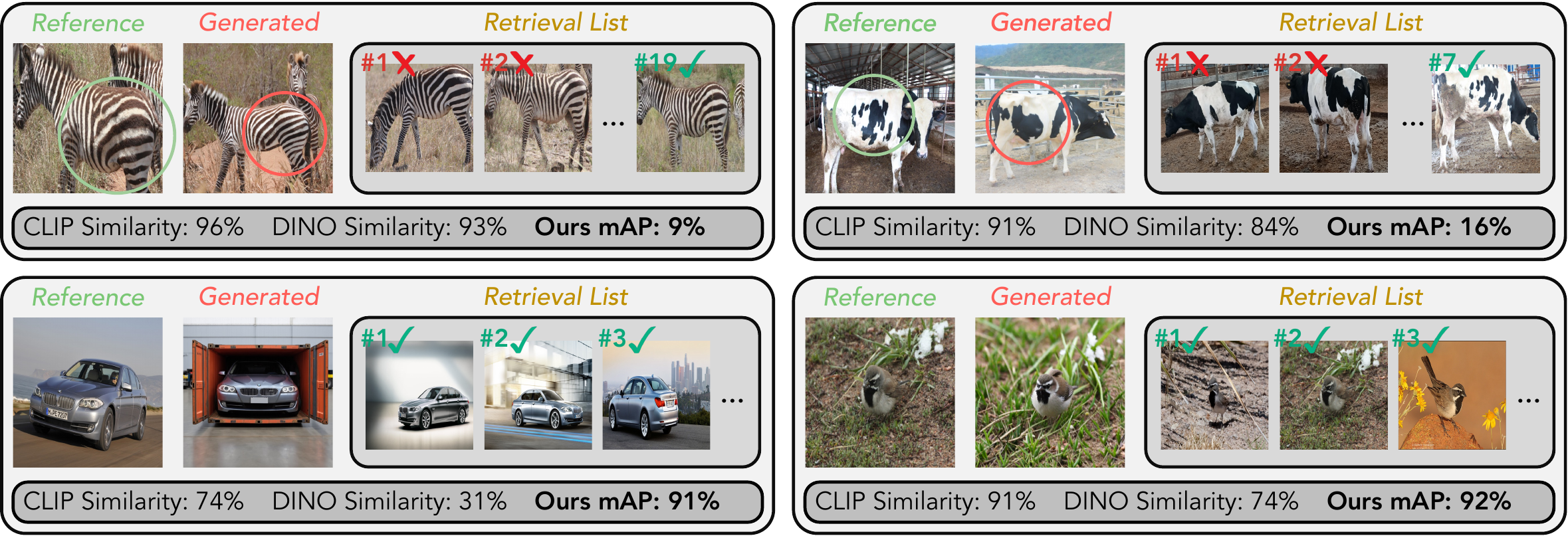}
    \caption{\textbf{Qualitative comparisons} between pair-wise similarity scores and the gallery-based \Ours. Each group shows the reference image used in prompting generation, the generated image, and the retrieval list using the generated image as the query. Our evaluation protocol reveals the variations in subtle details and demonstrates robustness against context variations.}
    \label{fig:qualitative}
    \vspace{-4pt}
\end{figure*}

\subsection{Oracle Experiments}
We first consider real images used in similarity calculation and gallery probing as an oracle case to examine the effectiveness of different visual encoders. As shown in \autoref{tab:oracle}, CLIP and DINO yield high scores when calculating similarities on Re-ID and CUB datasets, where images share more similar backgrounds and contexts. Comparatively, more diverse backgrounds in Stanford Cars lead to lower similarities for these two models. BioCLIP and SP-Cars are trained with fine-grained objectives, and thus demonstrate robust performance. MiewID does not yield high similarity scores, partly because it faithfully relies on details to recognize identities. After fine-grained training, all three models have much higher mAP scores than CLIP and DINO when evaluated with retrieval. These results support that the specialized models are more capable of distinguishing identity-related variations and are more suitable for the gallery-based evaluation.

\subsection{Main Results}
We then look into the generated images of various personalization models in \autoref{fig:main-result} (detailed values listed in the Supplementary Material). We vary two factors in this main experiment: the evaluation protocol (pair-wise similarity \vs gallery-based retrieval) and the encoder (general-purpose CLIP/DINO \vs specialized models).

\paragraph{Evaluation protocol}
We show the results of pair-wise similarities and gallery-based mAP scores in the first and second rows, respectively. CLIP scores are the most broadly adopted metric for personalization models, as the model possesses alignment between images and textual semantics. The CLIP similarity scores suggest that the generated images across all the models have equally achieved 80\% preservation of the reference. However, we show that by introducing galleries with visually similar identities, the CLIP mAP scores demonstrate substantial degradation. DINO shows similarity variations across models, but performance drops when negative samples are involved in the evaluation.
Compared with the popularly adopted concept preservation metrics, \Ours reveals substantial identity drift across popular personalization models.

\paragraph{Specialized models}
While the introduction of visually similar negative samples offers more realistic comparisons, the general-purpose visual encoders may not be able to capture fine-grained identity-specific details, and thus restrict the evaluation quality. We further adopt specialized models in the evaluation, which assign more attention to those discriminative details. Such an enhancement calibrates the ``equally poor'' performances of CLIP and DINO mAP scores and faithfully uncovers variations across different models.
Overall, DreamBooth~\cite{ruiz2023dreambooth} yields the best capability in preserving the reference identity.

\subsection{Qualitative Examples}
We use \autoref{fig:qualitative} to intuitively demonstrate the effectiveness of the gallery-based \Ours in evaluating identity preservation of personalization models.

In the first row, we show the generation results of Re-ID images. In the zebra case, the generated image has a very similar object layout and background to the reference, leading to almost perfect CLIP and DINO similarities. In fact, the generated image loses the distinctive stripes of the reference, and via retrieval, our protocol correctly assigns a low identity preservation score. In the cow case, while it is obvious for humans to tell the difference between the generated image and the reference, the pair-wise similarity still yields high scores. 

We also show results for more general category-level datasets in the second row. Both the generated car and bird images successfully retain the original identity. However, for the car case, the context differences obscure the feature extraction of DINO, leading to a low similarity. Interfered with by such a large background variation, our proposed protocol correctly focuses on the discriminative details and assigns a high score for the generation. These qualitative examples clearly demonstrate the robustness of \Ours encountering similar or diverse contexts.
We provide detailed qualitative comparison across images generated by the evaluated methods in the Supplementary Material.

\begin{figure}[t]
    \centering
    \includegraphics[width=0.49\linewidth]{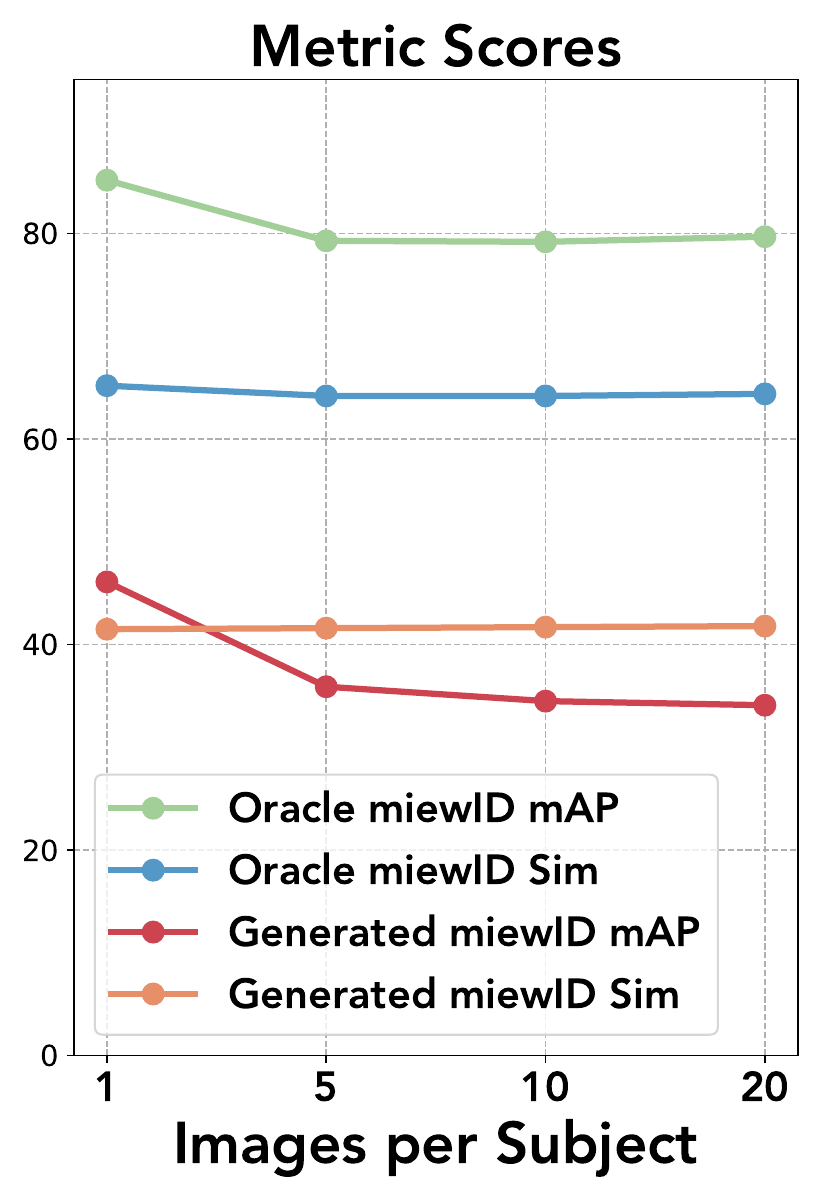}
    \includegraphics[width=0.49\linewidth]{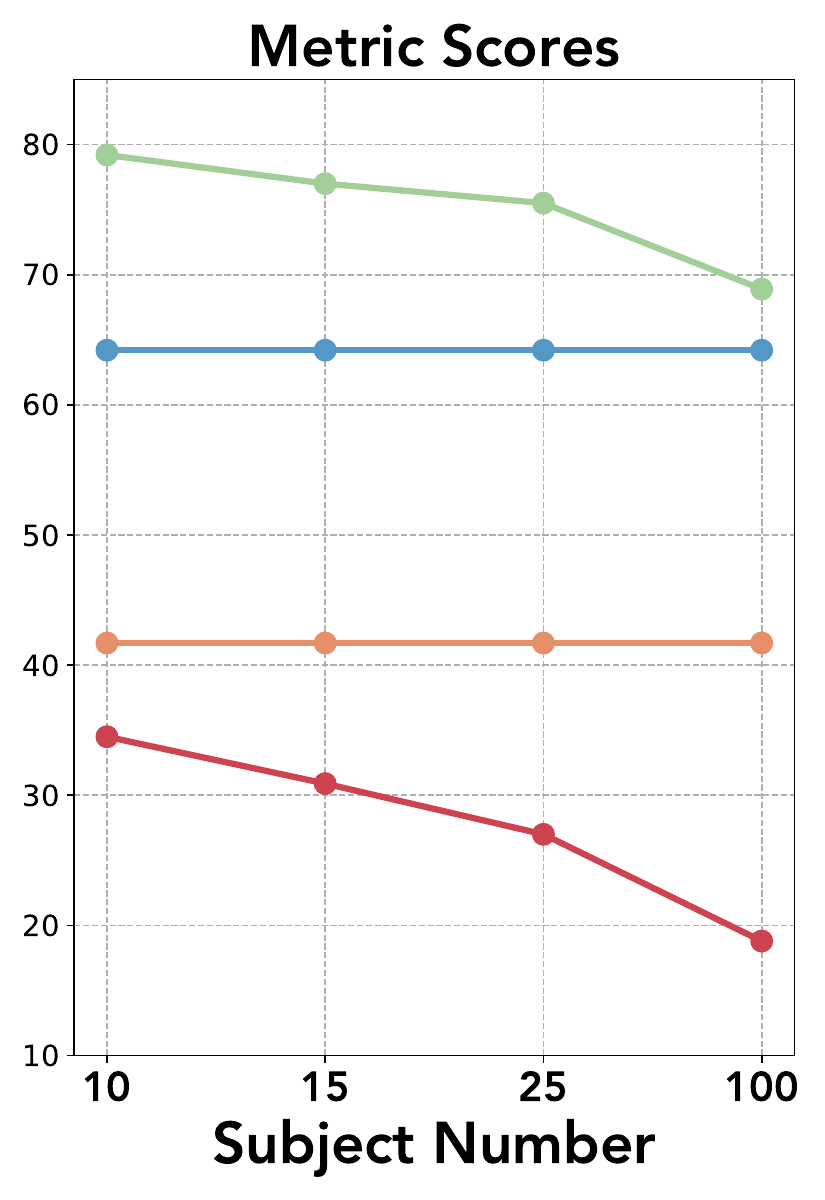}
    \caption{\textbf{Ablation study on the gallery construction}. We vary the image number per subject (left) and the subject number (right) in the gallery. ``Sim'' refers to Similarity. The generated performance is averaged over all adopted personalization models. Adding more subjects does not change pair-wise similarities, as the extra subjects are not involved in the calculation.}
    \label{fig:abl-gallery-size}
    \vspace{-4pt}
\end{figure}

\begin{table}[t]
    \centering
    \small
    \begin{tabular}{c|cccc}
    \toprule
       \multirow{2}{*}{Sampling}  & \multicolumn{2}{c}{Oracle} & \multicolumn{2}{c}{Generated} \\
       & mAP & Sim & mAP & Sim \\
    \midrule
        Random & $79.8$ & $64.3$ & $34.8$ & $41.6$ \\
        k-means & $78.6$ & $63.9$ & $34.4$ & $41.5$ \\
        Ours & $79.2$ & $64.2$ & $34.5$ & $41.7$ \\
    \bottomrule
    \end{tabular}
    \caption{\textbf{Ablation study on gallery sampling}. Re-ID datasets and the MiewID model are used in this experiment. The generated performance is averaged over all adopted personalization models. Different sampling brings a limited impact on the performances.}
    \label{tab:abl-gallery-sampling}
    \vspace{-6pt}
\end{table}

\subsection{Gallery Ablation Study}\label{sec:abl-gallery}
The key difference between the proposed protocol and pair-wise similarities is the introduction of the gallery. We therefore ablate the gallery design and discuss the principles of selecting appropriate gallery images.

We first report the effect of varying images per subject and the subject number in \autoref{fig:abl-gallery-size}. 
The experiments are conducted with real images for calculating similarity and probing galleries. Note that in this experiment, we take the similarity between all the corresponding gallery and generated images for a subject and then use the average of that. More images per subject introduces more views; thus enhancing the overall diversity of the gallery. Increasing the number from 1 to 5 increases the difficulty of retrieving the correct identities, while further increasing it to 20 only brings a moderate influence. Increasing the subject number alone doesn't introduce more information for each identity, but only raises the benchmark difficulty, leading to lower mAP scores. As the pair-wise similarity only involves the reference identity, adding subjects does not change the similarity scores. Considering both computational cost and task difficulty, we use ten subjects, each with ten images, for each of our adopted datasets.

We then ablate different sampling strategies to construct galleries from the original datasets in \autoref{tab:abl-gallery-sampling}. We aim to build galleries that have diverse views for each subject. The results show that both random sampling and k-means yield similar results to our adopted galleries, indicating that they lead to nearly equal diversity and ensure that the galleries avoid bias toward specific image contexts if possible.

\section{Discussions}\label{sec:discussion}

\begin{figure}
    \centering
    \includegraphics[width=0.9\linewidth]{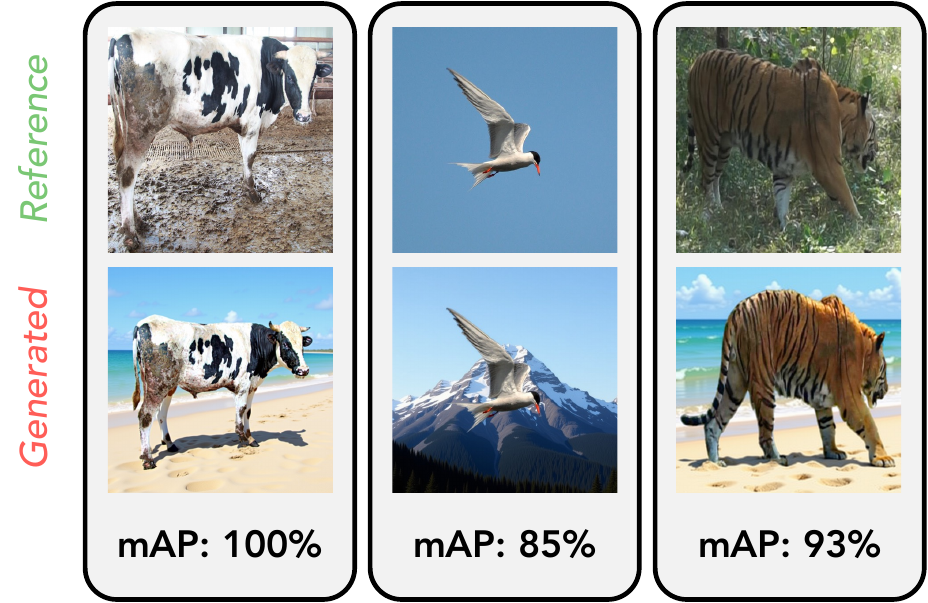}
    \vspace{-4pt}
    \caption{\textbf{Copy-pasting} objects can receive almost perfect identity preservation scores with the retrieval evaluation. The effectiveness of such generations requires more comprehensive examinations and assessments, such as instruction following. }
    \label{fig:copy-paste}
    \vspace{-8pt}
\end{figure}

\subsection{Pasting Effects}
A further consideration is the influence of copy-paste behavior on identity preservation evaluation. A personalization model can achieve strong ranking performance by reproducing large portions of the reference image rather than learning a robust representation of the identity. As shown in \autoref{fig:copy-paste}, when the personalization model performs image-editing style generation, the resulting images can get almost perfect scores.
In such cases, the high retrieval scores reflect memorization rather than generative preservation. Therefore, we state that ranking-based identity scores should not be used in isolation. Instead, they should be interpreted together with complementary measures and qualitative diagnostics that reveal overfitting and copy-paste artifacts, such as instruction following capabilities. 

\begin{figure}[t]
    \centering
    \includegraphics[width=\linewidth]{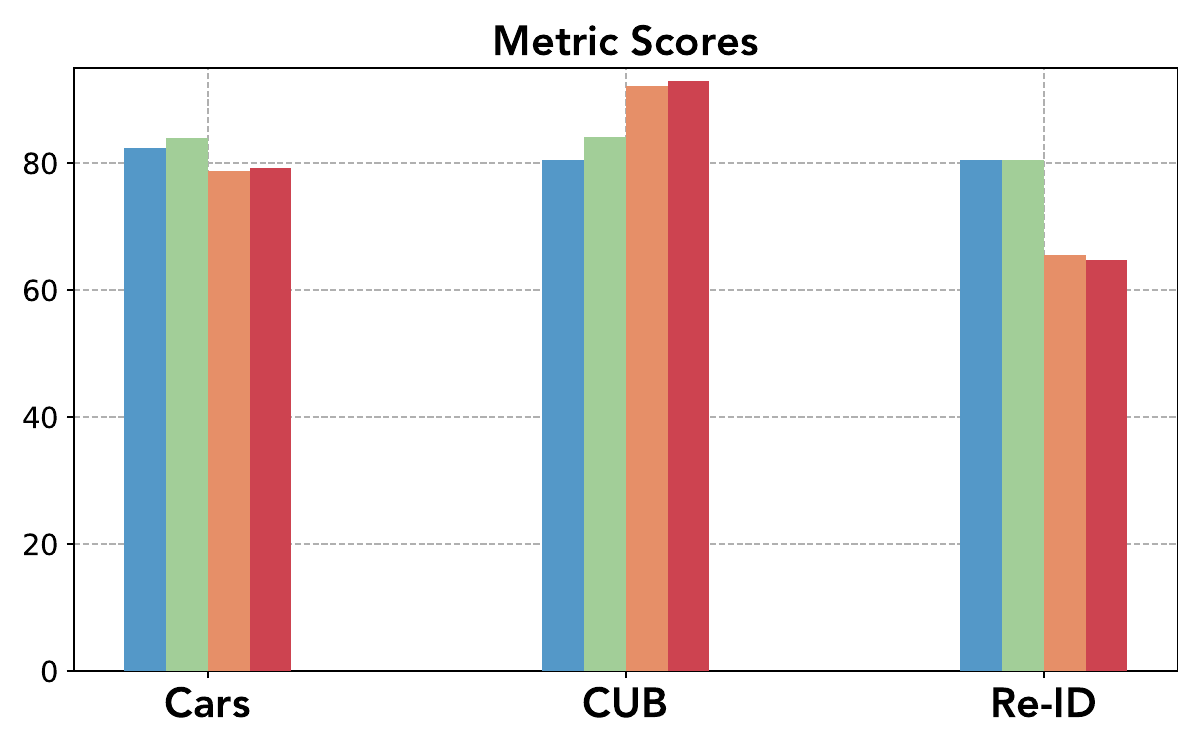}
    \includegraphics[width=0.75\linewidth]{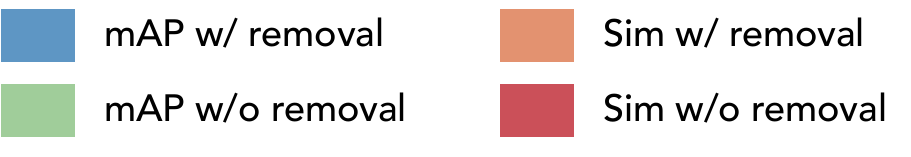}
    \caption{\textbf{The influence of background removal} on the oracle model performances. The small performance differences between w/ and w/o removal validate that the specialized models are robust against background variations.}
    \label{fig:background-removal}
\end{figure}

\subsection{Background Removal}\label{sec:background-removal}
Removing backgrounds has been proven to provide a better focus on the target objects~\cite{kotar2023these}. We compare the oracle model performance of removing the backgrounds or not in \autoref{fig:background-removal}. Specialized models are used in this experiment. We show that the overall mAP and similarity scores are only slightly influenced by removing or keeping backgrounds. It demonstrates that the adopted specialized models are robust to such background variations, with more focus on identity-specific details. Therefore, for simplicity and reproducibility, we retain backgrounds in our benchmarks.

\subsection{Retrieval or Classification}\label{sec:retrieval-or-classification}

Given a fine-grained dataset, the identity preservation can be evaluated by both retrieval and classification accuracy. 
While classification-based evaluation can verify whether a generated image is assigned to the correct identity, it offers a relatively coarse, binary signal. However, we care not only about predicting the correct identity, but also about how strongly the generated sample remains separated from visually similar alternatives. More concretely, we would want discriminative details instead of generic characteristics to be presented in the generated image. 
Retrieval-based evaluation provides a richer perspective. Ranking all candidate identities reveals whether the generated image is distributed in a neighborhood dominated by its target identity, or whether it is easily confused with near-duplicate individuals~\cite{chen2022deep}. 
If a method partially preserves identity but loses key discriminative details, the mAP score can accurately reflect the potentially reduced rank. Conversely, a classification-based metric would treat such cases the same as perfectly preserved identities, as long as the top-1 prediction is the correct identity. Therefore, retrieval is more aligned with the requirements in subject-driven personalization. 
We provide a more detailed discussion about the advantage of retrieval in the Supplementary Material.

\subsection{Broader Applications}
Beyond evaluating personalized generative models, the proposed gallery-based protocol has direct implications for the use of synthetic data in training downstream recognition systems~\cite{tremblay2018training,nikolenko2021synthetic,barbosa2018looking}. When synthetic images are used to augment or even replace real data in fine-grained classification or Re-ID, the retention of category/individual-specific features will be the key factor to data effectiveness. A generator that produces visually similar but identity-ambiguous samples can increase diversity while injecting label noise. In contrast, a generator that scores well under our retrieval-based evaluation is more likely to provide synthetic data that supports downstream training. This suggests a broader role for the identity preservation metric as a diagnostic tool for selecting and tuning generative pipelines before large-scale data synthesis. Moreover, it can also serve as a criterion to discard failed synthetic samples.  

\subsection{Ethical Statement}
We recognize that identity-sensitive evaluation raises important ethical considerations, particularly when applied to data that may contain biometric or personally identifiable information. Our protocol is designed to be instantiated on datasets where identities correspond to non-human subjects and does not require collecting additional information beyond the labels provided by the datasets. Nevertheless, the same methodology could be extended to human identities. We therefore strongly discourage deploying this evaluation on sensitive human biometric data without careful review by institutional ethics boards. We recommend that users avoid combining our protocol with data sources that may harm vulnerable populations and prioritize transparency about what identities are being modeled and for what purpose. 

\section{Conclusions}
We propose \Ours, a gallery-based protocol for evaluating identity preservation of personalized generation models. Building on the insight that pair-wise similarity fails to emphasize identity-specific details, we introduce galleries consisting of visually similar identities. Incorporating models specialized for fine-grained tasks, we reveal that popular personalization models still suffer from substantial identity drift in subject-driven generation scenarios. Through this protocol, we emphasize the importance of fine-grained comparison in evaluating identity preservation, and call for method development tailored for preserving identities in practical and diverse generation applications.

\section*{Acknowledgment}
This research is supported in part by grants from the
National Science Foundation (OAC-2118240, HDR Institute:Imageomics). The authors are grateful for the generous
support of the computational resources from the Ohio Supercomputer Center.


{
    \small
    \bibliographystyle{ieeenat_fullname}
    \bibliography{main}
}
\clearpage
\clearpage
\setcounter{page}{1}
\appendix

\section*{Appendix}

\noindent
The appendix is organized as follows:
\begin{itemize}
    \item \autoref{sec:app-implementation} details the hyperparameters and inference settings adopted in our experiments.
    \item \autoref{sec:app-retrieval-or-classification} discusses the choice of retrieval instead of classification in this paper.
    \item \autoref{sec:app-qualitative-comparison} shows the images generated by all the methods that we evaluate.
    \item \autoref{sec:app-numerical} lists the detailed numerical results of all the evaluated methods on our benchmark.
    \item \autoref{sec:app-prompt-following} discusses the metrics for prompt following that supplement our retrieval evaluation.
\end{itemize}

\begin{table}[h]
    \vskip 2pt
    \centering
    \small
    \caption{\textbf{DreamBooth fine-tuning hyperparameters.} BS: batch size, LR: learning rate, PPW: prior preservation weight}
    \begin{tabular}{cccccc}
    \toprule
        \textbf{Method} & \textbf{T2I model} & \textbf{BS} & \textbf{LR} & \textbf{Steps} & \textbf{PPW}\\
    \midrule
        DreamBooth & SD v3.0 & $1$ & $5$e-$6$ & $800$ & $0.5$ \\
    \bottomrule
    \end{tabular}
    \label{tab:finetune}
\end{table}

\begin{table}[h]
    \centering
    \small
    \caption{\textbf{Inference hyperparameters.} GS: guidance scale, IS: inference steps}
    \begin{tabular}{lcc}
    \toprule
        \textbf{Method} & \textbf{GS} & \textbf{IS}\\
    \midrule
        DreamBooth & $5.0$ & $30$\\
        OminiControl & $3.5$ & $30$ \\
        Flux Kontext & $5.0$ & $30$ \\
        Emu2 & $3.0$ & $50$ \\
        Diptych & $3.5$ & $30$ \\
        IP-Adapter & $5.0$ & $50$ \\
        IP-Adapter-Plus & $5.0$ & $50$ \\
    \bottomrule
    \end{tabular}
    \label{tab:inference}
\end{table}

\section{Implementation Details}\label{sec:app-implementation}
We adjust the hyperparameters of each method within common ranges based on qualitative testing and quantitative performances. 
Grounded by user-community preference, DreamBooth is the only method we evaluate that needs fine-tuning. The fine-tuning configurations are listed in \autoref{tab:finetune}. 
For inference, both IP-Adapter and IP-Adapter-Plus used an image condition weight factor of 0.6, similar to that in previous works~\cite{peng2025dreambench++,shin2025large}. We list the hyperparameters used for each method in \autoref{tab:inference}. In general, near-ideal settings were chosen for each method.
IP-Adapter, IP-Adapter-Plus, Diptych, and Flux Kontext use the negative prompt of: \textit{low quality, cropped, deformed, text, ugly, blurry, zoomed in, cartoony, unrealistic, lowres, not showing whole subject, animated}.

In \autoref{sec:abl-gallery}, we ablate on the gallery selection process for each subject in the Re-ID dataset, where randomly sampling corresponding subject gallery images and using k-means are compared. When creating the k-means gallery, the subject's candidate images are passed through MiewID~\cite{otarashvili2024multispecies} to obtain feature embeddings. K-means is then applied to acquire 10 clusters. The image closest to each cluster centroid is selected to construct the gallery.

\clearpage

\noindent
\begin{minipage}{\textwidth}
\begin{center}
    \centering
    \includegraphics[width=1\textwidth]{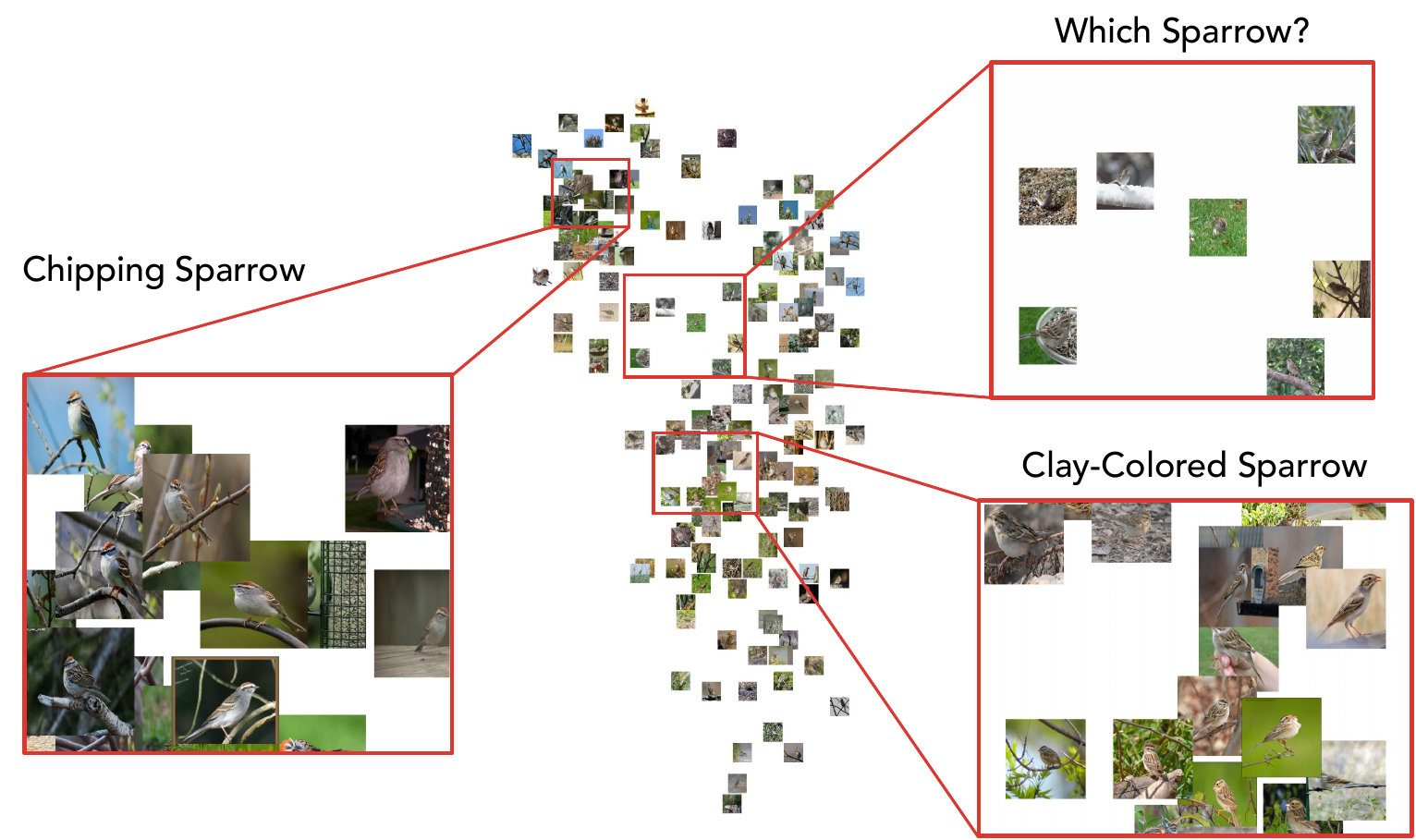}
    \captionof{figure}{\textbf{Distribution visualization of CUB sparrow species}. The images in the central regions of each species usually show more pronounced traits. By contrast, those in the boundary regions are hard to distinguish due to a less representative display of features. }
    \label{fig:tsne}
    \vskip 10pt
\end{center}
\end{minipage}

\section{Retrieval or Classification?}\label{sec:app-retrieval-or-classification}
We demonstrate in \autoref{sec:retrieval-or-classification} that while classification also validates the identity, it assigns samples at class boundaries with the same score as those at class centroid regions. We further illustrate this in \autoref{fig:tsne}, where t-SNE visualization is presented using the original CUB images. 
The cap of the chipping sparrow is reddish brown, while that of the clay-colored sparrow has black stripes. 
They can also be distinguished by the dark eyeline, which extends through the eye to the bill for chipping sparrows. 
Comparatively, the images distributed in the central region of the chipping sparrow cluster show more pronounced traits, and so is true for the clay-colored sparrow images. However, at the boundary region, these features are not clearly displayed. 

In personalized generation, users would like instance-unique details to be well-presented. Thereby, they can easily tell that the generated image contains \textit{their} pets or belongings. By contrast, images with vague details should be avoided. However, classification does not punish these vague cases as long as the generated image is correctly classified. When retrieval is applied instead, the generated images may retrieve real images of other identities, degrading the mAP score. In this way, retrieval helps emphasize discriminative details that are critical in personalized generation applications compared with solely classification. 

\clearpage

\noindent
\begin{minipage}{\textwidth}
\begin{center}
    \centering
    \includegraphics[width=\textwidth]{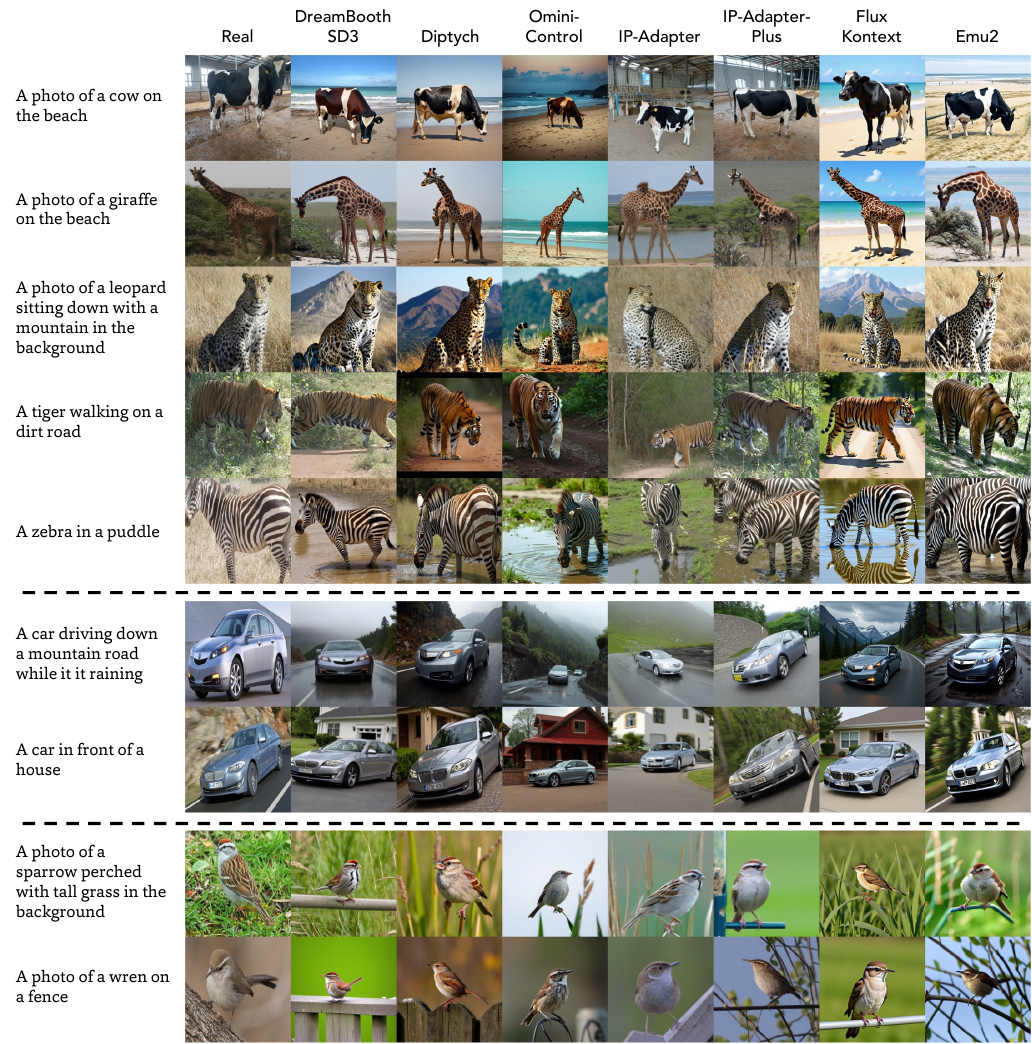}
    \captionof{figure}{\textbf{Qualitative comparison between different methods on \Ours}. Overall, DreamBooth achieves the best identity preservation and instruction following capabilities.}
    \label{fig:generation-example}
\end{center}
\end{minipage}
\clearpage
\section{Qualitative Comparison}\label{sec:app-qualitative-comparison}

We present the example images generated by different methods across adopted datasets in \autoref{fig:generation-example}. Aligned with our quantitative results in \autoref{fig:main-result}, DreamBooth~\cite{ruiz2023dreambooth} tends to yield overall better capabilities in preserving fine-grained subject identity. Particularly, with the cow and zebra as references in \autoref{fig:generation-example}, DreamBooth generates images that better represent the identity-specific spots and patterns. 

Comparing the results of our retrieval evaluation with the other metrics in \autoref{fig:main-result}, we observe the following insights.
First, CLIP mainly emphasizes the generic semantics contained in the images. CLIP embeddings generally yield high similarities between the reference and generated images, especially in the animal Re-ID scenarios. 
DINO tends to uncover more variations across methods, as the model is trained to focus on relations between different image patches. Second, Emu2~\cite{sun2024generative} and IP-Adapter-Plus~\cite{ye2023ip} are assigned high scores across benchmarks by both CLIP and DINO similarities. From qualitative examples in \autoref{fig:generation-example}, we find that these two methods sometimes cannot follow the personalization instructions. For example, IP-Adapter-Plus often generates images with a nearly identical background to the reference. While with a similar layout, the generation still cannot retain discriminative details, such as patterns of animals. These details are not reflected by either CLIP or DINO similarities. 

At a broader view of detail with CUB and Stanford Cars, the performances of different methods shift between datasets. With cars, DreamBooth still seems to objectively best capture the original identity in its generation, while with birds, IP-Adapter-Plus is the model that captures most identity details. We hypothesize that DreamBooth excels at capturing finer details after deliberate fine-tuning. By contrast, methods like IP-Adapter and Emu2 are better at leveraging semantics accumulated during the generative pre-training. Therefore, when presented with bird species, they have a better capability in recovering the correct traits of the specific bird. Especially when the generated poses are different from the reference, these methods produce better results than DreamBooth.

Cross-examining the fine-grained generated output from Re-ID in \autoref{fig:generation-example} with previous metrics, a possible explanation for the shift in method rankings is discovered. Methods like IP-Adapter-Plus and Emu2 strongly perform in the Re-ID evaluation on prior similarity metrics, but qualitative examination of the generated output of these methods reveals strong semantic similarities with the real subject, but lacking in identity-specific detail. Corroborated by findings in \cite{peng2025dreambench++}, CLIP and DINO seem to prefer non-identity-defining features such as shape, color, and style. This reaffirms the need for data-specialized models to be used when measuring the preservation of identity, as these models are optimized to focus on identity-specific details.

\begin{table*}[!h]
\centering
\small
\caption{\textbf{Leopard ReID Results}}
\label{tab:ReID_leopard}
\begin{tabular}{l|ccc|ccc|c|c}
\toprule
\multirow{2}{*}{\textbf{Method}} & \textbf{CLIP-I} & \textbf{DINO} & \textbf{MiewID} & \textbf{CLIP-I} & \textbf{DINO} & \textbf{MiewID} & \multirow{2}{*}{\textbf{DB++}} &\multirow{2}{*}{\textbf{CLIP-T}}\\
& \textbf{Sim}   & \textbf{Sim}  & \textbf{Sim} & \textbf{mAP}   & \textbf{mAP}  & \textbf{mAP} & \\
\midrule
Diptych\cite{shin2025large} & $0.825$ & $0.779$ & $0.409$ & $0.238$ & $0.287$ & $0.465$ & $0.762$ & $0.318$ \\
DreamBooth\cite{ruiz2023dreambooth} & $0.845$ & $0.762$ & $0.393$ & $0.282$ & $0.384$ & $0.669$ & $0.680$ & $0.321$ \\
Emu2\cite{sun2024generative} & $0.885$ & $0.801$ & $0.420$ & $0.272$ & $0.351$ & $0.493$ & $0.885$ & $0.302$ \\
Flux Kontext\cite{labs2025flux} & $0.743$ & $0.617$ & $0.313$ & $0.201$ & $0.226$ & $0.385$ & $0.500$ & $0.333$ \\
IP Adapter\cite{ye2023ip} & $0.882$ & $0.768$ & $0.299$ & $0.254$ & $0.268$ & $0.248$ & $0.790$ & $0.307$ \\
IP Adapter Plus\cite{ye2023ip} & $0.913$ & $0.825$ & $0.389$ & $0.307$ & $0.324$ & $0.429$ & $0.925$ & $0.294$ \\
OminiControl\cite{tan2025ominicontrol} & $0.775$ & $0.646$ & $0.272$ & $0.210$ & $0.208$ & $0.218$ & $0.560$ & $0.342$ \\
\bottomrule
\end{tabular}
\end{table*}

\begin{table*}[!h]
\centering
\small
\caption{\textbf{Cow ReID Results}}
\label{tab:ReID_cow}
\begin{tabular}{l|ccc|ccc|c|c}
\toprule
\multirow{2}{*}{\textbf{Method}} & \textbf{CLIP-I} & \textbf{DINO} & \textbf{MiewID} & \textbf{CLIP-I} & \textbf{DINO} & \textbf{MiewID} & \multirow{2}{*}{\textbf{DB++}} &\multirow{2}{*}{\textbf{CLIP-T}}\\
& \textbf{Sim}   & \textbf{Sim}  & \textbf{Sim} & \textbf{mAP}   & \textbf{mAP}  & \textbf{mAP} & \\
\midrule
Diptych\cite{shin2025large} & $0.746$ & $0.632$ & $0.461$ & $0.173$ & $0.225$ & $0.341$ & $0.697$ & $0.332$ \\
DreamBooth\cite{ruiz2023dreambooth} & $0.782$ & $0.643$ & $0.472$ & $0.201$ & $0.255$ & $0.483$ & $0.628$ & $0.327$ \\
Emu2\cite{sun2024generative} & $0.792$ & $0.710$ & $0.534$ & $0.230$ & $0.326$ & $0.380$ & $0.852$ & $0.305$ \\
Flux Kontext\cite{labs2025flux} & $0.721$ & $0.624$ & $0.510$ & $0.192$ & $0.292$ & $0.580$ & $0.665$ & $0.337$ \\
IP Adapter\cite{ye2023ip} & $0.864$ & $0.705$ & $0.463$ & $0.250$ & $0.193$ & $0.205$ & $0.602$ & $0.280$ \\
IP Adapter Plus\cite{ye2023ip} & $0.926$ & $0.808$ & $0.555$ & $0.353$ & $0.299$ & $0.324$ & $0.850$ & $0.254$ \\
OminiControl\cite{tan2025ominicontrol} & $0.677$ & $0.452$ & $0.349$ & $0.158$ & $0.162$ & $0.206$ & $0.465$ & $0.334$ \\
\bottomrule
\end{tabular}
\end{table*}

\begin{table*}[!h]
\centering
\small
\caption{\textbf{Zebra ReID Results}}
\label{tab:ReID_zebra}
\begin{tabular}{l|ccc|ccc|c|c}
\toprule
\multirow{2}{*}{\textbf{Method}} & \textbf{CLIP-I} & \textbf{DINO} & \textbf{MiewID} & \textbf{CLIP-I} & \textbf{DINO} & \textbf{MiewID} & \multirow{2}{*}{\textbf{DB++}} &\multirow{2}{*}{\textbf{CLIP-T}}\\
& \textbf{Sim}   & \textbf{Sim}  & \textbf{Sim} & \textbf{mAP}   & \textbf{mAP}  & \textbf{mAP} & \\
\midrule
Diptych\cite{shin2025large} & $0.875$ & $0.831$ & $0.600$ & $0.175$ & $0.182$ & $0.246$ & $0.877$ & $0.331$ \\
DreamBooth\cite{ruiz2023dreambooth} & $0.882$ & $0.818$ & $0.598$ & $0.192$ & $0.214$ & $0.406$ & $0.778$ & $0.339$ \\
Emu2\cite{sun2024generative} & $0.900$ & $0.835$ & $0.587$ & $0.181$ & $0.226$ & $0.202$ & $0.888$ & $0.317$ \\
Flux Kontext\cite{labs2025flux} & $0.824$ & $0.761$ & $0.584$ & $0.173$ & $0.194$ & $0.335$ & $0.683$ & $0.351$ \\
IP Adapter\cite{ye2023ip} & $0.912$ & $0.833$ & $0.558$ & $0.170$ & $0.165$ & $0.160$ & $0.830$ & $0.315$ \\
IP Adapter Plus\cite{ye2023ip} & $0.936$ & $0.892$ & $0.607$ & $0.183$ & $0.198$ & $0.166$ & $0.952$ & $0.305$ \\
OminiControl\cite{tan2025ominicontrol} & $0.842$ & $0.727$ & $0.514$ & $0.176$ & $0.176$ & $0.154$ & $0.690$ & $0.344$ \\
\bottomrule
\end{tabular}
\end{table*}

\begin{table*}[!h]
\centering
\small
\caption{\textbf{Tiger ReID Results}}
\label{tab:ReID_tiger}
\begin{tabular}{l|ccc|ccc|c|c}
\toprule
\multirow{2}{*}{\textbf{Method}} & \textbf{CLIP-I} & \textbf{DINO} & \textbf{MiewID} & \textbf{CLIP-I} & \textbf{DINO} & \textbf{MiewID} & \multirow{2}{*}{\textbf{DB++}} &\multirow{2}{*}{\textbf{CLIP-T}}\\
& \textbf{Sim}   & \textbf{Sim}  & \textbf{Sim} & \textbf{mAP}   & \textbf{mAP}  & \textbf{mAP} & \\
\midrule
Diptych\cite{shin2025large} & $0.812$ & $0.788$ & $0.436$ & $0.180$ & $0.234$ & $0.418$ & $0.712$ & $0.327$ \\
DreamBooth\cite{ruiz2023dreambooth} & $0.863$ & $0.858$ & $0.511$ & $0.252$ & $0.447$ & $0.720$ & $0.770$ & $0.321$ \\
Emu2\cite{sun2024generative} & $0.868$ & $0.844$ & $0.473$ & $0.263$ & $0.336$ & $0.478$ & $0.880$ & $0.303$ \\
Flux Kontext\cite{labs2025flux} & $0.778$ & $0.739$ & $0.408$ & $0.160$ & $0.245$ & $0.435$ & $0.610$ & $0.339$ \\
IP Adapter\cite{ye2023ip} & $0.869$ & $0.785$ & $0.347$ & $0.250$ & $0.187$ & $0.201$ & $0.778$ & $0.302$ \\
IP Adapter Plus\cite{ye2023ip} & $0.897$ & $0.848$ & $0.489$ & $0.332$ & $0.269$ & $0.564$ & $0.945$ & $0.294$ \\
OminiControl\cite{tan2025ominicontrol} & $0.816$ & $0.701$ & $0.311$ & $0.163$ & $0.163$ & $0.201$ & $0.602$ & $0.342$ \\
\bottomrule
\end{tabular}
\end{table*}

\begin{table*}[!h]
\centering
\small
\caption{\textbf{Giraffe ReID Results}}
\label{tab:ReID_giraffe}
\begin{tabular}{l|ccc|ccc|c|c}
\toprule
\multirow{2}{*}{\textbf{Method}} & \textbf{CLIP-I} & \textbf{DINO} & \textbf{MiewID} & \textbf{CLIP-I} & \textbf{DINO} & \textbf{MiewID} & \multirow{2}{*}{\textbf{DB++}} &\multirow{2}{*}{\textbf{CLIP-T}}\\
& \textbf{Sim}   & \textbf{Sim}  & \textbf{Sim} & \textbf{mAP}   & \textbf{mAP}  & \textbf{mAP} & \\
\midrule
Diptych\cite{shin2025large} & $0.866$ & $0.739$ & $0.394$ & $0.212$ & $0.204$ & $0.257$ & $0.720$ & $0.322$ \\
DreamBooth\cite{ruiz2023dreambooth} & $0.866$ & $0.716$ & $0.384$ & $0.230$ & $0.289$ & $0.401$ & $0.655$ & $0.324$ \\
Emu2\cite{sun2024generative} & $0.906$ & $0.775$ & $0.436$ & $0.267$ & $0.264$ & $0.317$ & $0.868$ & $0.301$ \\
Flux Kontext\cite{labs2025flux} & $0.813$ & $0.685$ & $0.437$ & $0.211$ & $0.267$ & $0.470$ & $0.628$ & $0.334$ \\
IP Adapter\cite{ye2023ip} & $0.917$ & $0.747$ & $0.331$ & $0.206$ & $0.193$ & $0.185$ & $0.650$ & $0.309$ \\
IP Adapter Plus\cite{ye2023ip} & $0.937$ & $0.809$ & $0.416$ & $0.259$ & $0.265$ & $0.279$ & $0.893$ & $0.299$ \\
OminiControl\cite{tan2025ominicontrol} & $0.825$ & $0.603$ & $0.303$ & $0.174$ & $0.185$ & $0.194$ & $0.578$ & $0.334$ \\
\bottomrule
\end{tabular}
\end{table*}


\begin{table*}[!h]
\centering
\small
\caption{\textbf{CUB: Sparrow Results}}
\label{tab:CUB_sparrow}
\begin{tabular}{l|ccc|ccc|c|c}
\toprule
\multirow{2}{*}{\textbf{Method}} & \textbf{CLIP-I} & \textbf{DINO} & \textbf{BioCLIP} & \textbf{CLIP-I} & \textbf{DINO} & \textbf{BioCLIP} & \multirow{2}{*}{\textbf{DB++}} &\multirow{2}{*}{\textbf{CLIP-T}}\\
& \textbf{Sim}   & \textbf{Sim}  & \textbf{Sim} & \textbf{mAP}   & \textbf{mAP}  & \textbf{mAP} & \\
\midrule
Diptych\cite{shin2025large} & $0.771$ & $0.547$ & $0.771$ & $0.307$ & $0.469$ & $0.524$ & $0.615$ & $0.336$ \\
DreamBooth\cite{ruiz2023dreambooth} & $0.776$ & $0.531$ & $0.805$ & $0.353$ & $0.542$ & $0.677$ & $0.585$ & $0.315$ \\
Emu2\cite{sun2024generative} & $0.868$ & $0.608$ & $0.828$ & $0.462$ & $0.663$ & $0.695$ & $0.825$ & $0.294$ \\
Flux Kontext\cite{labs2025flux} & $0.721$ & $0.404$ & $0.736$ & $0.323$ & $0.566$ & $0.628$ & $0.530$ & $0.321$ \\
IP Adapter\cite{ye2023ip} & $0.858$ & $0.626$ & $0.805$ & $0.448$ & $0.472$ & $0.519$ & $0.695$ & $0.309$ \\
IP Adapter Plus\cite{ye2023ip} & $0.898$ & $0.712$ & $0.866$ & $0.498$ & $0.597$ & $0.708$ & $0.925$ & $0.291$ \\
OminiControl\cite{tan2025ominicontrol} & $0.755$ & $0.500$ & $0.777$ & $0.330$ & $0.574$ & $0.675$ & $0.620$ & $0.327$ \\
\bottomrule
\end{tabular}
\end{table*}

\begin{table*}[!h]
\centering
\small
\caption{\textbf{CUB: Tern Results}}
\label{tab:CUB_tern}
\begin{tabular}{l|ccc|ccc|c|c}
\toprule
\multirow{2}{*}{\textbf{Method}} & \textbf{CLIP-I} & \textbf{DINO} & \textbf{BioCLIP} & \textbf{CLIP-I} & \textbf{DINO} & \textbf{BioCLIP} & \multirow{2}{*}{\textbf{DB++}} &\multirow{2}{*}{\textbf{CLIP-T}}\\
& \textbf{Sim}   & \textbf{Sim}  & \textbf{Sim} & \textbf{mAP}   & \textbf{mAP}  & \textbf{mAP} & \\
\midrule
Diptych\cite{shin2025large} & $0.821$ & $0.727$ & $0.840$ & $0.287$ & $0.321$ & $0.446$ & $0.725$ & $0.336$ \\
DreamBooth\cite{ruiz2023dreambooth} & $0.802$ & $0.684$ & $0.853$ & $0.320$ & $0.383$ & $0.570$ & $0.700$ & $0.349$ \\
Emu2\cite{sun2024generative} & $0.879$ & $0.774$ & $0.895$ & $0.366$ & $0.459$ & $0.610$ & $0.885$ & $0.315$ \\
Flux Kontext\cite{labs2025flux} & $0.755$ & $0.543$ & $0.798$ & $0.295$ & $0.351$ & $0.588$ & $0.605$ & $0.334$ \\
IP Adapter\cite{ye2023ip} & $0.872$ & $0.750$ & $0.868$ & $0.317$ & $0.314$ & $0.405$ & $0.800$ & $0.330$ \\
IP Adapter Plus\cite{ye2023ip} & $0.913$ & $0.821$ & $0.911$ & $0.370$ & $0.405$ & $0.609$ & $0.950$ & $0.320$ \\
OminiControl\cite{tan2025ominicontrol} & $0.782$ & $0.601$ & $0.818$ & $0.297$ & $0.354$ & $0.480$ & $0.675$ & $0.333$ \\
\bottomrule
\end{tabular}
\end{table*}

\begin{table*}[!h]
\centering
\small
\caption{\textbf{CUB: Wren Results}}
\label{tab:CUB_wren}
\begin{tabular}{l|ccc|ccc|c|c}
\toprule
\multirow{2}{*}{\textbf{Method}} & \textbf{CLIP-I} & \textbf{DINO} & \textbf{BioCLIP} & \textbf{CLIP-I} & \textbf{DINO} & \textbf{BioCLIP} & \multirow{2}{*}{\textbf{DB++}} &\multirow{2}{*}{\textbf{CLIP-T}}\\
& \textbf{Sim}   & \textbf{Sim}  & \textbf{Sim} & \textbf{mAP}   & \textbf{mAP}  & \textbf{mAP} & \\
\midrule
Diptych\cite{shin2025large} & $0.781$ & $0.428$ & $0.713$ & $0.299$ & $0.432$ & $0.477$ & $0.625$ & $0.329$ \\
DreamBooth\cite{ruiz2023dreambooth} & $0.771$ & $0.482$ & $0.748$ & $0.324$ & $0.456$ & $0.509$ & $0.520$ & $0.321$ \\
Emu2\cite{sun2024generative} & $0.876$ & $0.642$ & $0.826$ & $0.464$ & $0.676$ & $0.720$ & $0.875$ & $0.303$ \\
Flux Kontext\cite{labs2025flux} & $0.721$ & $0.265$ & $0.658$ & $0.296$ & $0.409$ & $0.437$ & $0.380$ & $0.309$ \\
IP Adapter\cite{ye2023ip} & $0.863$ & $0.635$ & $0.809$ & $0.408$ & $0.474$ & $0.592$ & $0.765$ & $0.332$ \\
IP Adapter Plus\cite{ye2023ip} & $0.905$ & $0.773$ & $0.902$ & $0.480$ & $0.635$ & $0.736$ & $0.955$ & $0.306$ \\
OminiControl\cite{tan2025ominicontrol} & $0.766$ & $0.385$ & $0.715$ & $0.333$ & $0.456$ & $0.516$ & $0.550$ & $0.313$ \\
\bottomrule
\end{tabular}
\end{table*}


\begin{table*}[!h]
\centering
\small
\caption{\textbf{Stanford Cars Results}}
\label{tab:Cars}
\begin{tabular}{l|ccc|ccc|c|c}
\toprule
\multirow{2}{*}{\textbf{Method}} & \textbf{CLIP-I} & \textbf{DINO} & \textbf{SP-Cars} & \textbf{CLIP-I} & \textbf{DINO} & \textbf{SP-Cars} & \multirow{2}{*}{\textbf{DB++}} &\multirow{2}{*}{\textbf{CLIP-T}}\\
& \textbf{Sim}   & \textbf{Sim}  & \textbf{Sim} & \textbf{mAP}   & \textbf{mAP}  & \textbf{mAP} & \\
\midrule
Diptych\cite{shin2025large} & $0.752$ & $0.596$ & $0.751$ & $0.298$ & $0.206$ & $0.664$ & $0.873$ & $0.281$ \\
DreamBooth\cite{ruiz2023dreambooth} & $0.734$ & $0.418$ & $0.770$ & $0.335$ & $0.211$ & $0.762$ & $0.687$ & $0.275$ \\
Emu2\cite{sun2024generative} & $0.836$ & $0.754$ & $0.758$ & $0.339$ & $0.194$ & $0.670$ & $0.918$ & $0.236$ \\
Flux Kontext\cite{labs2025flux} & $0.724$ & $0.440$ & $0.745$ & $0.305$ & $0.195$ & $0.719$ & $0.670$ & $0.266$ \\
IP Adapter\cite{ye2023ip} & $0.737$ & $0.448$ & $0.519$ & $0.237$ & $0.157$ & $0.428$ & $0.568$ & $0.272$ \\
IP Adapter Plus\cite{ye2023ip} & $0.783$ & $0.628$ & $0.459$ & $0.235$ & $0.173$ & $0.353$ & $0.420$ & $0.231$ \\
OminiControl\cite{tan2025ominicontrol} & $0.624$ & $0.304$ & $0.412$ & $0.170$ & $0.134$ & $0.337$ & $0.530$ & $0.291$ \\
\bottomrule
\end{tabular}
\end{table*}

\begin{table*}[!h]
\centering
\small
\caption{\textbf{Background Removal Oracle Results}}
\label{tab:background_remove_full}
\resizebox{\textwidth}{!}{
\begin{tabular}{cc|ccccc|ccccc}
\toprule
\multirow{2}{*}{\textbf{Dataset}} & \multirow{2}{*}{\textbf{Background}} &  \textbf{CLIP} & \textbf{DINO} & \textbf{MiewID} & \textbf{BioCLIP} & \textbf{SP-Cars} & \textbf{CLIP} & \textbf{DINO} & \textbf{MiewID} & \textbf{BioCLIP} & \textbf{SP-Cars} \\
&  & \textbf{Sim}   & \textbf{Sim}  & \textbf{Sim}    & \textbf{Sim}     & \textbf{Sim} & \textbf{mAP}   & \textbf{mAP}  & \textbf{mAP}    & \textbf{mAP}     & \textbf{mAP} \\
\midrule
\multirow{2}{*}{Re-ID} & Kept & $0.925$ & $0.843$ & $0.647$ &  - & - & $0.485$ & $0.516$ & $0.803$ & - & - \\
& Removed & $0.936$ & $0.829$ & $0.656$ & - & - & $0.422$ & $0.476$ & $0.805$ & - & - \\
\midrule
\multirow{2}{*}{CUB} & Kept & $0.837$ & $0.726$ & - & $0.930$ & - & $0.520$ & $0.674$ & - & $0.842$ & - \\
& Removed & $0.879$ & $0.717$ & - & $0.921$ & - & $0.498$ & $0.639$ & - & $0.805$ & - \\
\midrule
\multirow{2}{*}{Car} & Kept & $0.782$ & $0.434$ & - & - & $0.792$ & $0.396$ & $0.228$ & - & - & $0.839$ \\
& Removed & $0.851$ & $0.474$ & - & - & $0.787$ & $0.414$ & $0.235$ & - & - & $0.824$ \\
\bottomrule
\end{tabular}
}
\end{table*}

\section{Detailed Numerical Results}\label{sec:app-numerical}
\autoref{tab:ReID_leopard} to \autoref{tab:Cars} present detailed numeric results for each metric, dataset, and method that are previously summarized in \autoref{fig:main-result}. \autoref{tab:background_remove_full} gives numeric results for background removal in \autoref{sec:background-removal}. 

Reflected in class-specific results, the class of subject being generated can have an impact on the ranking and performance of the personalization methods. Within animal Re-ID evaluation, preserving tiger identity in generation seems to be a simpler task, whereas the giraffe generation is much harder, reflected by numeric results in \autoref{tab:ReID_leopard} to \autoref{tab:ReID_giraffe}. Intuitively, this makes sense as some classes, such as giraffe, can have more intricate and numerous identity-defining features than others, such as tiger, which have modest identity-defining features. Similar findings can also be found within the CUB evaluations (\autoref{tab:CUB_sparrow} to \autoref{tab:CUB_wren}).

The Re-ID evaluation was designed to offer a more fine-grained and rigorous evaluation compared to using CUB and Stanford Cars. As expected, Re-ID evaluations in \Ours yield overall much lower scores compared to CUB and Stanford Cars.

Background removal has been proven to improve focus on target subjects~\cite{kotar2023these}, specifically for measuring image similarity. In \autoref{sec:background-removal} and \autoref{fig:background-removal}, we demonstrated that background removal has negligible impacts when using a specialized model. In \autoref{tab:background_remove_full}, we demonstrate that using background removal for classical metrics such as DINO and CLIP-I scores does minimally improve their similarity scores but negligibly impacts their retrieval ability. Ultimately, using a specialized model without background removal offers a more competitive yet efficient method to evaluate subject identity preservation.

\section{Metric for Prompt Following}\label{sec:app-prompt-following}

CLIP-T has been the de facto metric used in previous works to measure method prompt following abilities. CLIP-T measures the cosine similarity between the CLIP embeddings of the text generation prompt and the generated image, aligning with the CLIP~\cite{radford2021learning} training objective. Dreambench++~\cite{peng2025dreambench++} uses GPT as an alternative, while admitting CLIP-T and their proposed metric offer similar performance. However, factoring in compute expense, CLIP-T still offers the most efficient way to measure prompt following capabilities. Thus, we only evaluate prompt following using CLIP-T and report the results in~\autoref{tab:ReID_leopard}-\autoref{tab:Cars}. In general, \Ours should be used in conjunction with CLIP-T in order to provide a more robust evaluation of personalization methods.

\end{document}